%% file: SISC_main.tex
\documentclass[review]{siamart}

\input{SISC_shared}

\ifpdf
\hypersetup{
  pdftitle={\TheTitle},
  pdfauthor={\TheAuthors}
}
\fi

\begin{document}

\maketitle

\begin{abstract}
 Gradient-enhanced Kriging (GE-Kriging) is a well-established surrogate modelling technique for approximating expensive computational models. However, it tends to get impractical for high-dimensional problems due to the size of the inherent correlation matrix and the associated high-dimensional hyper-parameter tuning problem. To address these issues, a new method, called sliced GE-Kriging (SGE-Kriging), is developed in this paper for reducing both the size of the correlation matrix and the number of hyper-parameters.  We first split the training sample set into multiple slices, 
  and invoke Bayes' theorem to approximate the full likelihood function
  via a sliced likelihood function, in which multiple small correlation matrices are utilized to describe the correlation of the sample set rather than one large one. Then, we replace the original high-dimensional hyper-parameter tuning problem with a low-dimensional counterpart by learning the relationship between the hyper-parameters and the derivative-based global sensitivity indices. The performance of SGE-Kriging is finally validated by means of numerical experiments with several benchmarks and a high-dimensional aerodynamic modeling problem. The results show that the SGE-Kriging model features an accuracy and robustness that is comparable to the standard one but comes at much less training costs. The benefits are most evident for high-dimensional problems with tens of variables.
\end{abstract}

\begin{keywords}
  Gradient-enhanced Kriging,  
  Surrogate modeling,
  Maximum likelihood estimation,
  Global sensitivity analysis,
  Hyper-parameter tuning
\end{keywords}

\begin{AMS}
 \href{https://mathscinet.ams.org/msc/msc2010.html?t=65D05}{65D05}
 \href{https://mathscinet.ams.org/msc/msc2010.html?t=60G15}{60G15}
\href{https://mathscinet.ams.org/msc/msc2010.html?t=60G15}{62G08}
 	 
\end{AMS}
%
%
%

\input{GE-Kriging_local_doc.tex}
\bibliographystyle{siamplain}
\bibliography{SISC_main}
\end{document}

%% file: SISC_shared.tex
\usepackage{lipsum}
\usepackage{amsfonts}
\usepackage{graphicx}
\usepackage{amsopn}

\usepackage{mathtools}
\usepackage{booktabs}

\usepackage{enumerate}

\usepackage{color}
\usepackage{changes}
\newcommand{\tc}{\textcolor{red}}

\usepackage{epstopdf}
\usepackage{algorithmic}

\newsiamremark{remark}{Remark}

\newcommand{\R}{\mathbb R}

\makeatletter
\renewcommand*\env@matrix[1][*\c@MaxMatrixCols c]{%
  \hskip -\arraycolsep
  \let\@ifnextchar\new@ifnextchar
  \array{#1}}
\makeatother

\newcount\Comments  
\newcommand{\kibitz}[2]{\ifnum\Comments=1\textcolor{#1}{#2}\fi}

\ifpdf
  \DeclareGraphicsExtensions{.eps,.pdf,.png,.jpg}
\else
  \DeclareGraphicsExtensions{.eps}
\fi

\newcommand{\TheTitle}{Sliced gradient-enhanced Kriging for high-dimensional function approximation} 
\newcommand{\TheAuthors}{Kai Cheng and Ralf Zimmermann}

\headers{Sliced gradient-enhanced Kriging}{\TheAuthors}

\title{{\TheTitle}} 

\author{
   Kai Cheng\thanks{Department of Mathematics and Computer Science, University of Southern Denmark (SDU) Odense,
	(\email{kai@sdu.dk})}
\and 
  Ralf Zimmermann\thanks{Department of Mathematics and Computer Science, University of Southern Denmark (SDU) Odense,
    (\email{zimmermann@imada.sdu.dk})}
}


%% file: GE-Kriging_local_doc.tex
\section{Introduction} 
 Surrogate models have been widely applied to alleviate the heavy computational burden in the field of design, control, optimization and uncertainty quantification, where expensive high-fidelity numerical simulation models are employed to study and simulate the underlying physical phenomena of various complex systems \cite{queipo2005surrogate}.
 A surrogate model, also known as a meta-model, is an interpolation or regression model based on sampled data that is to mimic the behavior of system responses with respect to selected input parameters, and it can be used to predict the response of the expensive simulation model at any untried point efficiently. In the past decades, various surrogate modelling techniques have been developed, including Kriging \cite{krige1951statistical,doi:10.1137/21M1432739,sacks1989design}, support vector regression \cite{vapnik1999nature,smola2004tutorial,cheng2021adaptive}, polynomial chaos expansion \cite{xiu2002wiener,blatman2011adaptive,cheng2018adaptive}, neural networks \cite{abiodun2018state,doi:10.1137/21M1397908} and so on.
 Among these approaches, Kriging has drawn the most attention in engineering applications due to its attractive statistical properties and its flexibility and extensibility.
 
 Kriging assumes that the system response is a realization of a Gaussian process  \cite{krige1951statistical,rasmussen2003gaussian}, and it can be fully described by the associated mean and covariance functions. The mean function is used to capture the global trend of a computational model. The covariance function, arguably the most important ingredient of Kriging, describes the spatial dependence between different sample points. Under the Gaussian process assumption, Kriging provides point-wise probabilistic prediction and error bars to account for the epistemic uncertainty.
 Hence, the Kriging method can be readily combined with adaptive sampling strategies \cite{jones1998efficient}.
 
 As with other surrogate modelling techniques, Kriging suffers from the so-called ``curse of dimensionality'', namely, the size of the sample set required to train an accurate Kriging model increases exponentially with the input dimension. To alleviate this problem, gradient-enhanced Kriging (GE-Kriging) has been developed. The idea is to  incorporate gradient information, obtained, say,  with adjoint methods \cite{anderson1999aerodynamic} or automatic differentiation \cite{neidinger2010introduction}, to enhance the accuracy of the Kriging model \cite{chung2002using,morris1993bayesian}.
 
 There are two basic versions of GE-Kriging in the literature, namely, indirect GE-Kriging and direct GE-Kriging. In indirect GE-Kriging \cite{liu2002gradient,liu2003development}, the gradient information is used to construct additional sample values and corresponding function values near the existing sample sites by a first-order Taylor's expansion. However, this method will  introduce additional numerical errors and is observed to result in ill-conditioned correlation matrices, which may undermine the performance of GE-Kriging. In contrast, by assuming that the system responses and its partial derivative functions are realizations of Gaussian processes, the direct GE-Kriging \cite{han2013improving,bouhlel2019gradient,han2017weighted,chen2020optimization} is constructed by absorbing the correlations between function values and the cross-correlations between function values and partial derivatives directly. Consequently, direct GE-Kriging is superior to the indirect approach in theory, and the corresponding correlation matrix is generally observed to be better conditioned \cite{han2017weighted,zimmermann2013maximum}. Therefore, direct GE-Kriging is investigated in current work. 
 
 Although GE-Kriging seems appealing to break the ``curse of dimensionality'', practical applications suggest that it suffers from two drawbacks in the context of high-dimensional problems. On the one hand, the correlation matrix of GE-Kriging is largely expanded, which makes it costly to deal with, especially when it comes to solving linear equation systems that involve the correlation matrix as the operator. On the other hand, the likelihood function of GE-Kriging is generally highly non-linear and contains multiple extrema, which makes it very challenging to tune the hyper-parameters of GE-Kriging by maximum likelihood estimation. As a consequence, training a GE-Kriging model can be extremely time-consuming for high-dimensional problems. This motivates researchers to improve the efficiency  of constructing GE-Kriging models. Bouhlel et al. \cite{bouhlel2019gradient} proposed to reduce the size of the correlation matrix of GE-Kriging as well as the number of hyper-parameters by utilizing a partial least-squares regression technique. Similarly,  Chen et al. \cite{chen2019screening}  suggested to reduce the correlation matrix in size by incorporating a part of the partial derivatives that is identified by a feature selection technique. In \cite{han2017weighted}, Han et al. proposed to train a series of submodels using only a part of the training sample set, and then to sum them up with appropriate weight coefficients to provide predictions.  
  
 In this paper, we develop a new method, called sliced GE-Kriging (SGE-Kriging) to improve the training efficiency of standard GE-Kriging for high-dimensional problems. As a first step, We utilize a derivative-based global sensitivity analysis method \cite{sobol2010derivative,kucherenko2016derivative} to explore the relative importance of every input variable with respect to the system response. Then, using Bayes' theorem, a simplified likelihood function is derived under the assumption that samples that are far away from each other are conditionally independent. In a second step, the number of hyper-parameters in the SGE-Kriging model is reduced by learning a preset functional relationship between the model hyper-parameters and the global sensitivity indices. Finally, several benchmarks are investigated to demonstrate the performance of SGE-Kriging. 
 
 The remainder of this paper is organized as follows. 
 In Section \ref{sec:GEK_basics}, we review the basic theory of GE-Kriging including the underlying assumptions, the key equations, the  hyper-parameter tuning approach and the associated correlation functions.  
 Then, the novel SGE-Kriging is introduced in Section \ref{sec:SGEK}. The derivative-based global sensitivity analysis method and the proposed sliced likelihood function are presented in Subsections \ref{sec:DevSense} and \ref{sec:slicedLF}, respectively, followed by the simplified hyper-parameter tuning method in Section \ref{sec:HPtuning}. We then compare the performance of the SGE-Kriging approach to the classical GE-Kriging in Section \ref{sec:numex} by means of several numerical benchmark examples of increasing sophistication and a high-dimensional aerodynamic modeling problem. Finally, we  summarize our observations and give conclusions in Section \ref{sec:conclusions}.

\section{A review of gradient-enhanced Kriging framework} 
\label{sec:GEK_basics}

In this section,  we provide an overview of standard GE-Kriging \cite{liu2003development,chung2002using}, followed by an introduction  of the existing hyper-parameter tuning technique and the correlation function models used in GE-Kriging. 

\subsection{ Gradient-enhanced Kriging model}
\label{sec:GEK}

 Consider a computational multiple input-single output process $y=g(\boldsymbol{x})$, where $\boldsymbol{x}=(x_1,\ldots,x_n )^\mathrm{T}$ is the input parameter vector, and $y$ is the scalar model response. GE-Kriging is concerned with approximating $g(\boldsymbol{x})$ by a model based on a set of observed samples $(\boldsymbol X,\boldsymbol y)$, where $\boldsymbol{X}$ is a matrix of $N$ sample sites $\boldsymbol{x}^{(i)}\in \R^{n}$, 
    and $\boldsymbol{y}$ is a response vector that contains the observed function values as well as the gradients at the sample sites, i. e.,
   \begin{equation*}
     \boldsymbol{X}=[\boldsymbol{x}^{(1)},\ldots,\boldsymbol{x}^{(N)}]^\mathrm{T}\in\R^{N\times n},
    \end{equation*}
   \begin{equation*}
   \boldsymbol{y}=[\boldsymbol{y}_0^\mathrm{T},\boldsymbol{y}_1^\mathrm{T},\ldots,
   \boldsymbol{y}_n^\mathrm{T}]
   ^\mathrm{T}\in\R^{(n+1)N},
  \end{equation*} 
  
 \noindent in which 
  \begin{equation*}
   \boldsymbol{y}_0 = [g(\boldsymbol{x}^{(1)}),g(\boldsymbol{x}^{(2)}),\ldots,g(\boldsymbol{x}^{(N)})]^\mathrm{T}\in\R^{N},
  \end{equation*}
  \begin{equation*}
   \boldsymbol{y}_k = \left[\frac{\partial g(\boldsymbol{x}^{(1)})}{\partial x_k},\frac{\partial g(\boldsymbol{x}^{(2)})}{\partial x_k},\ldots,\frac{\partial g(\boldsymbol{x}^{(N)})}{\partial x_k}\right]^\mathrm{T}\in\R^{N},\,\ k=1,\ldots,n.
  \end{equation*}
 
 
 In GE-Kriging, the computational model  $y=g(\boldsymbol{x})$ is regarded as a realization of a spatial Gaussian process $Y(\boldsymbol{x})$ that is given by a regression model and an additive Gaussian error term. In the case of constant regression, this means that
   \begin{equation}
   Y(\boldsymbol{x})=\beta_0 + Z(\boldsymbol{x}),
   \label{assumption}
   \end{equation}
   where $\beta_0$ is the regression constant that represents the mean value of $ Y(\boldsymbol{x})$, and  $Z(\boldsymbol{x})$ is a stationary Gaussian process with zero mean and covariance function 
   \begin{equation*}
   \text{Cov}\left(Z(\boldsymbol{x}),Z(\hat{\boldsymbol{x}})\right)=\sigma^2R(\boldsymbol{x},\hat{\boldsymbol{x}};\boldsymbol{\theta}).
   \end{equation*}
   
  \noindent In the latter, $\sigma^2$ is the stationary variance of $Z(\boldsymbol{x})$, and $R(\boldsymbol{x},\hat{\boldsymbol{x}};\boldsymbol{\theta})$ represents the spatial correlation between two points $\boldsymbol{x}$ and $\hat{\boldsymbol{x}}$ with the correlation lengths in the various coordinate directions being controlled by the hyper-parameter vector $\boldsymbol{\theta}$. 
 
 The partial derivative functions $\frac{\partial Z(\boldsymbol{x})}{\partial x_k}$ $(k=1,\ldots,n)$ of $Z(\boldsymbol{x})$ are again Gaussian processes, and the cross-covariance functions between the different Gaussian processes are given by the partial derivatives of the correlation function, see, e.~g., \cite[\S 4.5]{Koehler96computerexperiments} or \cite[ \S 4.1]{williams2006gaussian}:
    \begin{equation*}
    \text{Cov}\left(\frac{\partial Z({\boldsymbol{x}})}{\partial x_k},Z(\hat{\boldsymbol{x}})\right)=\sigma^2\frac{\partial R(\boldsymbol{x},\hat{\boldsymbol{x}};\boldsymbol{\theta})}{\partial x_k},\,\ k=1,\ldots,n,
   \end{equation*}
   \begin{equation*}
    \text{Cov}\left(Z(\boldsymbol{x}),\frac{\partial Z(\hat{\boldsymbol{x}})}{\partial \hat{x}_l}\right)=\sigma^2\frac{\partial R(\boldsymbol{x},\hat{\boldsymbol{x}};\boldsymbol{\theta})}{\partial \hat{x}_l},\,\ l=1,\ldots,n,
   \end{equation*}
    \begin{equation*}
     \text{Cov}\left(\frac{\partial Z(\boldsymbol{x})}{\partial x_k},\frac{\partial Z(\hat{\boldsymbol{x}})}{\partial \hat{x}_l}\right)=\sigma^2\frac{\partial^{2} R(\boldsymbol{x},\hat{\boldsymbol{x}};\boldsymbol{\theta})}{{\partial x_k}{\partial \hat{x}_l}},\,\ k,l=1,\ldots,n.
    \end{equation*}
  Under the Guassian process assumption, the joint distribution of $Y(\boldsymbol{x})$ and $\boldsymbol{Y}=[Y(\boldsymbol{x}^{(1)}),\ldots,Y(\boldsymbol{x}^{(N)}),\frac{\partial Y(\boldsymbol{x}^{(1)})}{\partial x_1},\ldots,\frac{\partial Y(\boldsymbol{x}^{(N)})}{\partial x_1},\ldots,\frac{\partial Y(\boldsymbol{x}^{(1)})}{\partial x_n},\ldots,\frac{\partial Y(\boldsymbol{x}^{(N)})}{\partial x_n}]^\mathrm{T}$ at the sample sites is multi\-variate Gaussian \cite{le2013multi}, namely
  
     \begin{equation*}
     \left[\begin{array}{cc}Y(\boldsymbol{x}) \\ \boldsymbol{Y} \end{array}\right] \sim \mathcal {N}\left(\left[\begin{array}{cc} \beta_0 \\ \beta_0\boldsymbol{F} \end{array}\right],\sigma^2\left[\begin{array}{cc} 1 & r^\mathrm{T} (\boldsymbol{x})\\ r(\boldsymbol{x}) & \boldsymbol{R} \end{array} \right] \right),
    \end{equation*}
     where
    \begin{equation*}
    \boldsymbol{F}= [\underbrace {1,\ldots,1}_N,\underbrace {0,\ldots,0}_{nN}]^\mathrm{T}\in\R^{(n+1)N}
    \end{equation*}
    for constant regression. In the case of a general regression model, $\boldsymbol{F}$ is to be replaced with a regression design matrix, see \cite[\S 2.3]{zimmermann2013maximum}.
    
     The matrix 
    \begin{equation*}
     \boldsymbol{R}= \begin{bmatrix}
     \boldsymbol{R}_{0,0} & \partial \boldsymbol{R}_{0,1} & \cdots & \partial\boldsymbol{R}_{0,n}\\
     \partial\boldsymbol{R}_{1,0} & \partial^2\boldsymbol{R}_{1,1} & \cdots & \partial^2\boldsymbol{R}_{1,n} \\ 
     \vdots & \vdots & \ddots & \vdots\\
     \partial\boldsymbol{R}_{n,0} & \partial^2\boldsymbol{R}_{n,1} & \cdots & \partial^2\boldsymbol{R}_{n,n}
    \end{bmatrix}   \in\R^{(n+1)N\times(n+1)N}
    \end{equation*}
    
  \noindent represents the correlation matrix of $\boldsymbol{Y}$, and its elements are given by
     \begin{eqnarray*}
     \boldsymbol{R}_{0,0}
     :=& R(\boldsymbol{x}^{(i)},{\boldsymbol{x}^{(j)}};\boldsymbol{\theta}),
     & i,j=1,\ldots,N,
     \\
     \partial\boldsymbol{R}_{0,k}=\partial\boldsymbol{R}_{k,0}^\mathrm{T}
     :=& \frac{\partial R(\boldsymbol{x}^{(i)},\boldsymbol{x}^{(j)};\boldsymbol{\theta})}{\partial {x^{(j)}_k}}, 
     & i,j=1,\ldots,N;k=1,\ldots,n,
    \\
    \partial^2\boldsymbol{R}_{k,l}
    :=&\frac{\partial^{2} R(\boldsymbol{x}^{(i)},\boldsymbol{x}^{(j)};\boldsymbol{\theta})}{{\partial {x^{(i)}_k}}{\partial {x^{(j)}_l}}},
    & i,j=1,\ldots,N;k,l=1,\ldots,n.
    \end{eqnarray*}
    
    Moreover, $r(\boldsymbol{x})=[r_0(\boldsymbol{x})^\mathrm{T},\partial r_1(\boldsymbol{x})^\mathrm{T},\ldots,\partial r_n(\boldsymbol{x})^\mathrm{T}]^\mathrm{T}\in\R^{(n+1)N}$ is the correlation vector between $Y(\boldsymbol{x})$ and $\boldsymbol{Y}$, in which
     \begin{eqnarray*}
     r_{0,i}(\boldsymbol{x}):= &R(\boldsymbol{x}^{(i)},{\boldsymbol{x}};\boldsymbol{\theta}),
     &i=1,\ldots,N,
    \\
     \partial r_{k,i}(\boldsymbol{x}):= 
     &\frac{\partial R(\boldsymbol{x}^{(i)},{\boldsymbol{x}};\boldsymbol{\theta})}{\partial {x_k}},
     & i=1,\ldots,N;k=1,\ldots,n.
     \end{eqnarray*}
     
     Given the observation $\boldsymbol{Y}=\boldsymbol{y}$, 
     the posterior predictive distribution $\hat Y(\boldsymbol{x})$ of $Y(\boldsymbol{x})$ is still Gaussian, namely 
     \begin{equation}
      \hat Y(\boldsymbol{x}) \sim \mathcal {N}\left(\mu(\boldsymbol{x}), s^2(\boldsymbol{x})\right),
     \label{predictor}
     \end{equation}
    where
      \begin{equation}
       \mu(\boldsymbol{x}) = \beta_0 +r^\mathrm{T}(\boldsymbol{x})\boldsymbol{R}^{-1}(\boldsymbol{y}-\beta_0\boldsymbol{F})  ,
       \label{meanfunction}
     \end{equation}
     and
     \begin{equation}
     s^2(\boldsymbol{x}) = \sigma^2 \left(1-r^\mathrm{T}(\boldsymbol{x})\boldsymbol{R}^{-1}r(\boldsymbol{x})+\left(1-\boldsymbol{F}^\mathrm{T}\boldsymbol{R}^{-1}r(\boldsymbol{x})\right)^2/(\boldsymbol{F}^\mathrm{T}\boldsymbol{R}^{-1}\boldsymbol{F})\right).
     \label{variancefunction}
    \end{equation}
     
     The mean function $\mu(\boldsymbol{x})$ serves as the surrogate model, and the variance function $s^2(\boldsymbol{x})$ measures the prediction uncertainty at the location $\boldsymbol{x}$. 
     
     \subsection{Maximum likelihood estimation}
     \label{sec:MLE}
     The performance of GE-Kriging depends on a number of parameters, namely, the regression constant $\beta_0$, the process variance $\sigma^2$ and the hyper-parameters $\boldsymbol{\theta}$. These parameters are generally tuned by the maximum likelihood estimation method. The likelihood function of GE-Kriging is
     \begin{equation}
     f(\boldsymbol{y})=\frac1{\sqrt{(2\pi\sigma^2)^{(n+1)N}[\rm det\boldsymbol R(\boldsymbol{\theta})}]}{\rm exp}\left(-\frac1{2\sigma^2}(\boldsymbol{y}-\beta_0\boldsymbol{F})^{\rm T}\boldsymbol{R}^{-1}(\boldsymbol{\theta})(\boldsymbol{y}-\beta_0\boldsymbol{F})\right).
     \label{originallikelihood}
    \end{equation}
     
    After taking the logarithm of Eq. (\ref{originallikelihood}), the optimal profile lines of $\beta_0$ and $\sigma^2$ depending on $\boldsymbol{\theta}$ can be derived analytically. This yields
     \begin{eqnarray}
      \beta_0(\boldsymbol{\theta}) &=& (\boldsymbol{F}^\mathrm{T}\boldsymbol{R}^{-1}(\boldsymbol{\theta})\boldsymbol{F})^{-1}\boldsymbol{F}^\mathrm{T}\boldsymbol{R}^{-1}(\boldsymbol{\theta})\boldsymbol{y},
     \label{beta} \\
     \sigma^2(\boldsymbol{\theta}) &=& \frac1{(n+1)N}(\boldsymbol{y}-\beta_0\boldsymbol{F})^{\rm T}\boldsymbol{R}^{-1}(\boldsymbol{\theta})(\boldsymbol{y}-\beta_0\boldsymbol{F}),
    \label{sigma2}
    \end{eqnarray}
    see, e.g, \cite {han2013improving}.
    There is no closed-form solutions for the optimal hyper-parameters $\boldsymbol{\theta}=[\theta_1,\ldots,\theta_n]^{\rm T}\in\R^{n}$ , and one has to use numerical optimization algorithms to determine their values. Substituting the optimal values of $\sigma^2(\boldsymbol{\theta})$ into the likelihood function in Eq. (\ref{originallikelihood}), we are left with minimizing the following $\boldsymbol{\theta}$-dependent likelihood function
    \begin{equation}
     \ell(\boldsymbol{\theta})=(n+1)N\ln \sigma^2(\boldsymbol{\theta}) + \ln\left[{\rm det\boldsymbol{R}(\boldsymbol{\theta})}\right].
   \label{reducedlikelihood}
   \end{equation}

    Note that when dealing with noisy data, one can assume that the noise of both the model response and its partial derivative is a Gaussian random variable with zero mean and constant variance $\lambda$. We can thus adapt the GE-Kriging (interpolation model) to a regression model by adding $\lambda$ to the diagonal element of the correlation matrix of standard GE-Kriging in Eq. \eqref{reducedlikelihood}, namely, $\boldsymbol{R}(\boldsymbol{\theta})+\lambda\boldsymbol{I}$. The noise variance can be determined together with $\boldsymbol{\theta}$ by minimizing the likelihood in Eq. \eqref{reducedlikelihood}.
   In the geostatistic literature, this is known as adding a ``nugget'' to the model.

    The likelihood function in Eq. \eqref{reducedlikelihood} is generally highly nonlinear and multimodal, and thus a global optimization algorithm is required to determine the hyper-parameters. To numerically evaluate this function, one needs to deal with a large, possibly ill-conditioned correlation matrix  $\boldsymbol{R}\in\R^{(n+1)N\times(n+1)N}$. When a Cholesky factorization is used to decompose $\boldsymbol{R}$, the corresponding computational costs are $\mathcal{O}((n+1)^3N^3)$, which tends to be very time-consuming for large number of samples $N$ and a large spatial dimension $n$. Moreover, the computations tend to be unstable.
    
    \subsection{Correlation function}
    \label{sec:CorrFun}
     In GE-Kriging, any twice continuously differentiable correlation function can be used. In general, a multivariate correlation function can be obtained via the tensor product of univariate ones, namely
     \begin{equation}
     R(\boldsymbol{x},\hat{\boldsymbol{x}};\boldsymbol{\theta})=\prod_{k=1}^nR(|x_k-\hat{x}_k|;\theta_k),
     \label{correlationfunction}
     \end{equation}
   where the inputs $\boldsymbol{x},\hat{\boldsymbol{x}}$ are $n$-vectors and $R(|x_k-\hat{x}_k|;\theta_k)$ denotes the one-dimensional spatial correlation function. 
   
    Although the Gaussian correlation function is popular in Kriging, it is known to lead to exceptionally ill-conditioned correlation matrices
    \cite{zimmermann2015}. In contrast, the cubic spline and biquadratic spline  correlation functions have been demonstrated to be comparably well-conditioned \cite{han2017weighted,rosenbaum2013efficient}. Based on our own numerical experiments (see the discussion in Section 4.1) and the literature \cite{rosenbaum2013efficient}, the biquadratic splines generally yield smoother and simpler likelihood functions when compared to the cubic spline correlation function. Therefore, the biquadratic spline correlation function is employed in this paper; the associated formulas are to be found in Appendix \ref{sec:biquadspline}. 
    
 \section{Sliced gradient-enhanced Kriging }
  \label{sec:SGEK}
 As mentioned previously, although the accuracy of a Kriging model can be improved significantly by incorporating gradient information, GE-Kriging suffers from two drawbacks for high-dimensional problems. First, the dimension of the correlation matrix of GE-Kriging is expanded from $N\times N$ to $N(n+1)\times N(n+1)$, when compared to Kriging. Second, as with Kriging, the likelihood function of GE-Kriging in Eq. \eqref{reducedlikelihood} generally contains multiple local minima in the high-dimensional space, and thus the hyper-parameter tuning process is numerically challenging. In this section, we will introduce 
 a modified approach, which we refer to as {\em sliced gradient-enhanced Kriging} (SGE-Kriging), with which we aim to address both the two aforementioned deficiencies of GE-Kriging simultaneously.

\subsection{Derivative-based global sensitivity analysis}
\label{sec:DevSense}
 Global sensitivity analysis aims at detecting a comparably small group of input variables that are deemed ``important'' from a large number candidate input variables.  This is achieved by quantifying the impact of the input variables on the model response \cite{cheng2020surrogate,wei2015variable,saltelli2008global}. Since we work under the assumption that gradient information is available, it is convenient to use the derivative-based global sensitivity analysis method in the work at hand. We employ the derivative-based global sensitivity index from \cite{sobol2010derivative,kucherenko2016derivative}, which is defined as
 \begin{equation}
 S_k = \int_{\R^n}\left(\frac{\partial g(\boldsymbol{x})}{\partial x_k}\right)^2\rho(\boldsymbol{x})d\boldsymbol{x},
 \label{eq:sensitivity_index}
  \end{equation}
 where $\rho(\boldsymbol{x})$ is the joint probability density function of $\boldsymbol{x}$. 
 The global sensitivity index in Eq. \eqref{eq:sensitivity_index} is originally developed in the field of uncertainty quantification, where it is used for ranking the  importance of the input variables, identifying the effective dimension of a computational model, and reducing the model complexity. In the current work, it is used to rank the importance of the input variables.
 Note that $S_k$ measures the variability of $g(\boldsymbol{x})$ in the $k$-th coordinate direction. For input variables with large sensitivity index, one can deduce that  $g(\boldsymbol{x})$ varies dramatically along the corresponding coordinate direction. In contrast, $g(\boldsymbol{x})$ is close to constant in the directions with small sensitivity index. Since GE-Kriging is generally constructed in a constrained box space, and physical parameter units should be normalized, it is recommended to scale the input parameter space to a hypercube $[0,1]^n$, and assumes that $\boldsymbol{x}$ are uniformly distributed random variables in the hypercube. Consequently, the derivative-based global sensitivity index can be estimated with the Monte Carlo method as
  \begin{equation}
     S_k \approx\hat S_k= \frac{1}{N}\sum_{i=1}^N \left(\frac{\partial g(\boldsymbol{x}^{(i)})}{\partial x_k}\right)^2,
  \end{equation}
 where $\boldsymbol{x}^{(i)}(i=1,...,N)$ are the observed sample sites. 

  The sensitivity indices provide us with useful information for tuning the hyper-parameters $\boldsymbol{\theta}=[\theta_1,\ldots,\theta_n]^{\rm T}$.  
  In fact, similar to the global sensitivity indices, the hyper-parameters $\theta_k (k=1,\ldots,n)$ that measure the correlation lengths in the various coordinate directions, can be interpreted as measuring how strongly the corresponding input variable $x_k(k=1,\ldots,n)$ affects the response of a GE-Kriging model \cite{bouhlel2019gradient,zhao2020efficient}. 
For an input variable with large sensitivity index, one can deduce that the hyper-parameter of GE-Kriging model on the corresponding direction tends to be large, and vice versa \cite{bouhlel2019gradient,zhao2020efficient}. 

\subsection{Sliced likelihood function}
\label{sec:slicedLF}
  In this subsection, we introduce the sliced likelihood function, which will be used to tune the hyper-parameter of SGE-Kriging. 
  With the sliced likelihood function, the correlation of the whole sample set can be described with multiple small matrices rather than a single big one.
  \begin{figure}[htp]
  \centering
  \includegraphics[width=8cm]{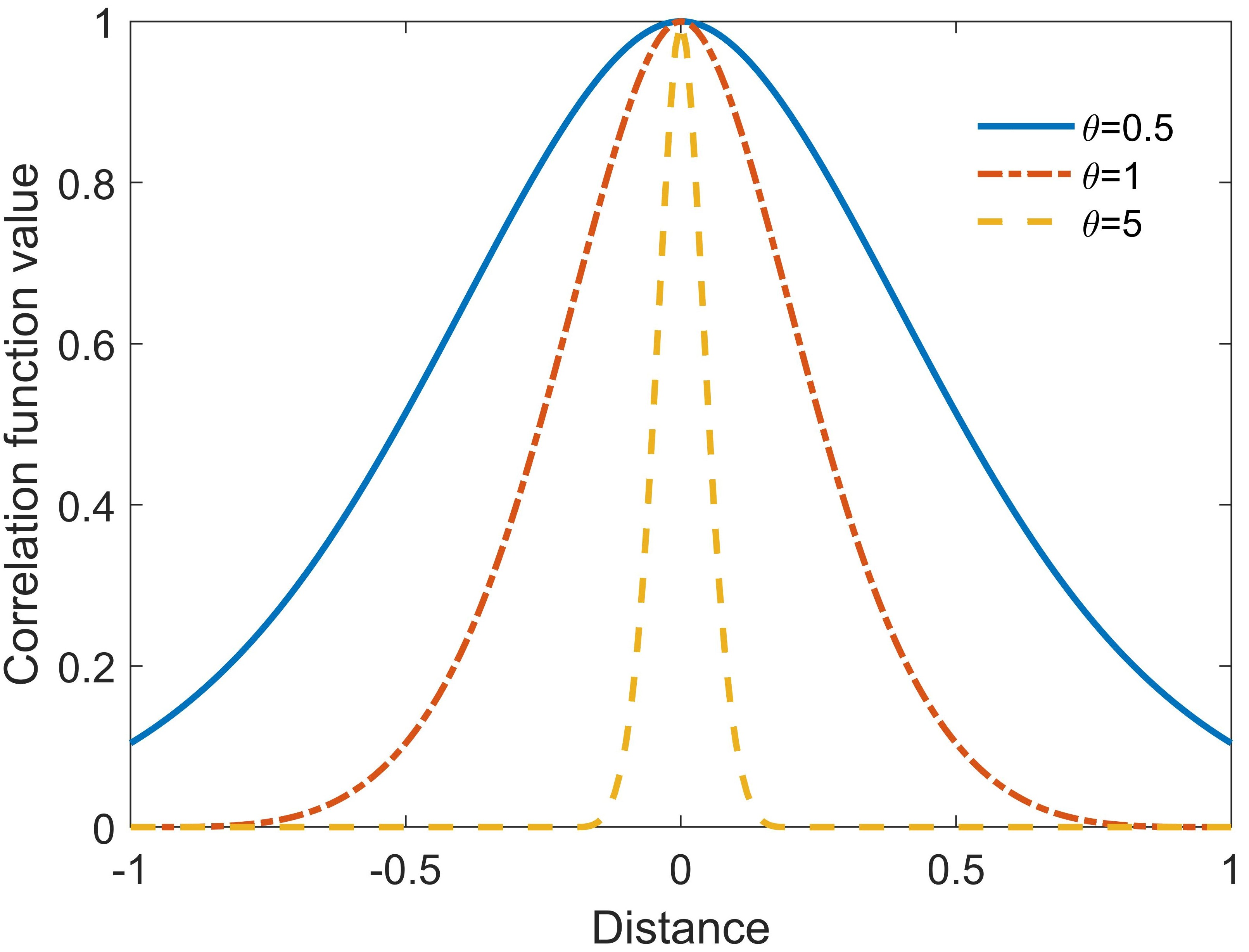}
  \caption{Biquadratic spline correlation function with different hyper-parameter $\theta$.}
  \label{biqudspline_figure}
  \end{figure}
  
\begin{figure}[htp]
  \centering
  \includegraphics[width=10cm]{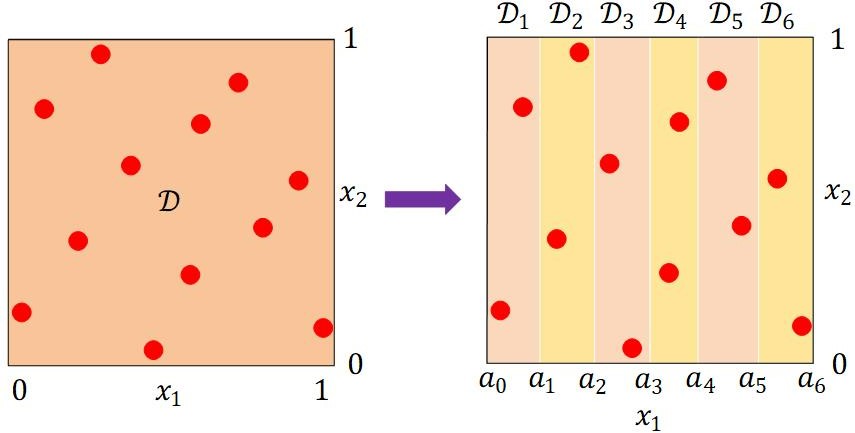}
  \caption{A two dimensional example of splitting the whole training sample set (left) into six slices (right) along the $x_1$-coordinate direction}
   \label{division}
  \end{figure}

   In Fig.~\ref{biqudspline_figure}, a one-dimensional biquadratic spline correlation function as detailed in Eq.~\eqref{biquadracticfunction} 
  is depicted.
  Obviously, along every direction $k=1,\ldots,n$, the sample correlation depends on the spatial distance $|x_k-\hat{x}_k|$ as well as on the value of the corresponding hyper-parameter $\theta_k$. The sample correlation is weak when the spatial distance is large or when the hyper-parameter is large. Therefore, we propose to neglect the correlation between samples that are far away from each other along direction with the largest hyper-parameter value. To this end, we divide the support [0,1] of the most important input variable $x_1$ (the input variable ranked with the largest sensitivity index) into $m$  successive and non-overlapping subintervals $\mathcal A_1=[a_0,a_1]$,\ldots,$\mathcal A_m=[a_{m-1},a_m]$, $a_0=0,a_m=1$, and the hypercube $[0,1]^n$ of the input parameter space is thus partitioned into $m$ slices $\mathcal D_1=[a_0,a_1]\times[0,1]^{n-1}$,\ldots,$\mathcal D_m=[a_{m-1},a_m]\times[0,1]^{n-1}$, as shown in Fig. \ref{division}. Consequently, the whole observed sample set $(\boldsymbol X,\boldsymbol y)$ is correspondingly splitted into $m$  non-overlapping subsets. Here we denote the subset of input sample sites and corresponding model response within each slices $\mathcal D_i$ as $\boldsymbol{ \tilde X}_i$ and $\boldsymbol{\tilde y}_i$ respectively.
  Thus, we have $\boldsymbol{X}=\boldsymbol{\tilde X}_1\cup\boldsymbol{\tilde X}_2\cup\dots\cup\boldsymbol{\tilde X}_m$, $\boldsymbol{y}=\boldsymbol{\tilde y}_1\cup\boldsymbol{\tilde y}_2\cup\dots\cup\boldsymbol{\tilde y}_m$. Each subset contains $N_i$ samples so that $\sum_{i=1}^m N_i=N$.
  
   Note that above partition strategy should guarantee that the number of samples within each slice should be as balanced as possible. To achieve this goal, we rank the $N$ samples of the most important input variable $x_1$ in index-ascending order as  ${x}^{(1)}_1,...,{x}^{(N)}_1$, and set $a_i=({x}^{(\sum_{j=1}^{i}{N_i})}_1+{x}^{(\sum_{j=1}^{i}{N_i}+1)}_1)/2, (i=1,...,m-1)$, where
  \begin{equation}
   N_i=\begin{cases}\lfloor N/m \rfloor+1, \,\ i=1,...,N_r,\\ \lfloor N/m \rfloor, \,\ \,\ \,\ \,\ i=N_r+1,...,m,
  \end{cases}
  \end{equation}
in which $N_r = N \,\ {\rm mod} \,\ m$ .
 
  Using Bayes' theorem, the likelihood function of GE-Kriging can be rewritten as
  \begin{equation}
    f(\boldsymbol{y})=f(\boldsymbol{\tilde y}_1)f(\boldsymbol{\tilde y}_2|\boldsymbol{\tilde y}_1)f(\boldsymbol{\tilde y}_3|\boldsymbol{\tilde y}_1,\boldsymbol{\tilde y}_2)\cdot\cdot\cdot f(\boldsymbol{\tilde y}_m|\boldsymbol{\tilde y}_1,\boldsymbol{\tilde y}_2,\ldots,\boldsymbol{\tilde y}_{m-1}).¨
   \label{truelikelihood}
  \end{equation}

  By assuming that the samples within non-adjacent slices are conditionally independent, namely, $(\boldsymbol{\tilde y}_{1},...,\boldsymbol{\tilde y}_{i-2})\bot\boldsymbol{\tilde y}_{i}|\boldsymbol{\tilde y}_{i-1}(i>2)$, we obtain 
 \begin{equation}
 \begin{aligned}
  f(\boldsymbol{\tilde y}_{i}|\boldsymbol{\tilde y}_{1},...,\boldsymbol{\tilde y}_{i-1})&=\frac{f(\boldsymbol{\tilde y}_{1},...,\boldsymbol{\tilde y}_{i-2},\boldsymbol{\tilde y}_{i}|\boldsymbol{\tilde y}_{i-1})f(\boldsymbol{\tilde y}_{i-1})}{f(\boldsymbol{\tilde y}_{1},...,\boldsymbol{\tilde y}_{i-1})}\\ &\approx \frac{f(\boldsymbol{\tilde y}_{1},...,\boldsymbol{\tilde y}_{i-2}|\boldsymbol{\tilde y}_{i-1})f(\boldsymbol{\tilde y}_{i}|\boldsymbol{\tilde y}_{i-1})f(\boldsymbol{\tilde y}_{i-1})}{f(\boldsymbol{\tilde y}_{1},...,\boldsymbol{\tilde y}_{i-1})} \\ &= f(\boldsymbol{\tilde y}_{i}|\boldsymbol{\tilde y}_{i-1}).
 \label{condindepedent}
 \end{aligned}
 \end{equation}
 
 With Eq. \eqref{condindepedent}, the original likelihood function $f(\boldsymbol{y})$ in  Eq. \eqref{truelikelihood} can be approximated by $\hat f(\boldsymbol{y})$ as
 \begin{equation}
 \begin{aligned}
 \hat f(\boldsymbol{y})&=f(\boldsymbol{\tilde y}_1)f(\boldsymbol{\tilde y}_2|\boldsymbol{\tilde y}_1)f(\boldsymbol{\tilde y}_3|\boldsymbol{\tilde y}_2)\cdot\cdot\cdot f(\boldsymbol{\tilde y}_m|\boldsymbol{\tilde y}_{m-1}).
 \label{cond_independent}
 \end{aligned}
 \end{equation}
  
 Note that the correlation between samples of two arbitrary non-adjacent slices, e.g., the $(i-1)$-th and $(i+1)$-th slices, are induced by samples of their common adjacent slice, i.e., the $i$-th slice, and the joint probability $f(\boldsymbol{\tilde y}_{i-1},\boldsymbol{\tilde y}_{i+1})$ can be approximately recovered as 
 \begin{equation}
 \hat f(\boldsymbol{\tilde y}_{i-1},\boldsymbol{\tilde y}_{i+1})=\int_{\R^{N_i(n+1)}} f(\boldsymbol{\tilde y}_{i-1}|\boldsymbol{\tilde y}_i)f(\boldsymbol{\tilde y}_{i+1}|\boldsymbol{\tilde y}_i)f(\boldsymbol{\tilde y}_i) d\boldsymbol{\tilde y}_i.
 \label{joint_distribution}
 \end{equation}
 
The likelihood function $\hat f(\boldsymbol{y})$ in  Eq. \eqref{cond_independent} can be further written as
 \begin{equation}
 \begin{aligned}
 \hat f(\boldsymbol{y})&=f(\boldsymbol{\tilde y}_1)f(\boldsymbol{\tilde y}_2|\boldsymbol{\tilde y}_1)f(\boldsymbol{\tilde y}_3|\boldsymbol{\tilde y}_2)\cdot\cdot\cdot f(\boldsymbol{\tilde y}_m|\boldsymbol{\tilde y}_{m-1})\\&= \frac{f(\boldsymbol{\tilde y}_1,\boldsymbol{\tilde y}_2)f(\boldsymbol{\tilde y}_2,\boldsymbol{\tilde y}_3)\cdot\cdot\cdot f(\boldsymbol{\tilde y}_{m-1},\boldsymbol{\tilde y}_m)}{f(\boldsymbol{\tilde y}_2)f(\boldsymbol{\tilde y}_3)\cdot\cdot\cdot f(\boldsymbol{\tilde y}_{m-1})}
 \\ &=\frac{\prod_{i=1}^{m-1}f(\boldsymbol{\tilde y}_i,\boldsymbol{\tilde y}_{i+1})}{\prod_{i=2}^{m-1}f(\boldsymbol{\tilde y}_i)}.
 \label{2appendant}
 \end{aligned}
 \end{equation}
  
 Since in Eq. \eqref{2appendant}, the correlation between samples from two adjacent slices are fully considered, we call the approximated likelihood function $\hat f(\boldsymbol{y})$ the {\em  2-appendant likelihood function}, and the corresponding GE-Kriging model by the name of {\em  2-appendant SGE-Kriging}. Note that this approach can be readily extended to incorporate the correlation of data from more than two slices. For example, the 3-appendant likelihood function is defined as
  \begin{equation}
  \begin{aligned}
    \hat f(\boldsymbol{y})&=f(\boldsymbol{\tilde y}_1)f(\boldsymbol{\tilde y}_2|\boldsymbol{\tilde y}_1)f(\boldsymbol{\tilde y}_3|\boldsymbol{\tilde y}_1,\boldsymbol{\tilde y}_2)f(\boldsymbol{\tilde y}_4|\boldsymbol{\tilde y}_2,\boldsymbol{\tilde y}_3)\cdot\cdot\cdot f(\boldsymbol{\tilde y}_m|\boldsymbol{\tilde y}_{m-2},\boldsymbol{\tilde y}_{m-1}) \\ &=\frac{f(\boldsymbol{\tilde y}_1,\boldsymbol{\tilde y}_2,\boldsymbol{\tilde y}_3)f(\boldsymbol{\tilde y}_2,\boldsymbol{\tilde y}_3,\boldsymbol{\tilde y}_4)\cdot\cdot\cdot f(\boldsymbol{\tilde y}_{m-2},\boldsymbol{\tilde y}_{m-1},\boldsymbol{\tilde y}_m)}{f(\boldsymbol{\tilde y}_2,\boldsymbol{\tilde y}_3)f(\boldsymbol{\tilde y}_3,\boldsymbol{\tilde y}_4)\cdot\cdot\cdot f(\boldsymbol{\tilde y}_{m-2},\boldsymbol{\tilde y}_{m-1})}
    \\ &=\frac{\prod_{i=1}^{m-2}f(\boldsymbol{\tilde y}_i,\boldsymbol{\tilde y}_{i+1},\boldsymbol{\tilde y}_{i+2})}{\prod_{i=2}^{m-2}f(\boldsymbol{\tilde y}_i,\boldsymbol{\tilde y}_{i+1})}.
    \label{3appendant} 
  \end{aligned}
  \end{equation}
  
   In theory, the 3-appendant likelihood function is closer to the original likelihood function since more sample correlations are retained. It is therefore expected to provide better approximations than the 2-appendant one from Eq. (\ref{2appendant}). However, according to our numerical experiments, the 2-appendant likelihood function already yields satisfactory approximation results for most tested problems, see the upcoming discussion in Section 4. Therefore, we proceed with the discussion of the 2-appendant likelihood function, and use this example to explain how to conduct the hyper-parameter tuning for SGE-Kriging. 
  
  Denoting $\boldsymbol{\tilde y}_{2*i}=[\boldsymbol{\tilde y}_{i},\boldsymbol{\tilde y}_{i+1}]^\mathrm{T}$, we have 
 \begin{eqnarray*}
f(\boldsymbol{\tilde y}_{2*i})&=&\frac{\exp\left(-\frac1{2\tilde\sigma^2}(\boldsymbol{\tilde y}_{2*i}-\tilde\beta_0\boldsymbol{\tilde F}_{2*i})^{\rm T}\boldsymbol{\tilde R}_{2*i}^{-1}(\boldsymbol{\tilde y}_{2*i}-\tilde\beta_0\boldsymbol{\tilde F}_{2*i})\right)}{\sqrt{(2\pi\tilde\sigma^2)^{\tilde N_{2*i}}[{\rm det}\boldsymbol{\tilde{R}}_{2*i}]}} ,\\
 f(\boldsymbol{\tilde y}_{ i})&=&\frac{{\rm exp}\left(-\frac1{2\tilde\sigma^2}(\boldsymbol{\tilde y}_{ i}-\tilde\beta_0\boldsymbol{\tilde F}_{i})^{\rm T}\boldsymbol{\tilde R}_{i,i}^{-1}(\boldsymbol{\tilde y}_{i}-\tilde\beta_0\boldsymbol{\tilde F}_{i})\right)}{\sqrt{(2\pi\tilde\sigma^2)^{\tilde N_{i}}[{\rm det}\boldsymbol{\tilde{R}}_{i,i}]}} ,
  \end{eqnarray*}
  where
 \begin{eqnarray*}
  \boldsymbol{\tilde F}_{2*i}&=& [\underbrace {1,\ldots,1}_{N_{i}},\underbrace {0,\ldots,0}_{nN_{i}},\underbrace {1,\ldots,1}_{N_{i+1}},\underbrace {0,\ldots,0}_{nN_{i+1}}]^\mathrm{T},\\ %
  \boldsymbol{\tilde F}_{i}&=& [\underbrace {1,\ldots,1}_{N_{i}},\underbrace {0,\ldots,0}_{nN_{i}}]^\mathrm{T},
 \end{eqnarray*}
 and the dimensions are
$\tilde N_{2*i}=(N_{i}+N_{i+1})(n+1)$,  and $\tilde N_{i}=N_{i}(n+1)$.
 The reduced correlation matrix is
\begin{equation*} \boldsymbol{{\tilde{R}}}_{2*i}= \left[\begin{array}{cc} \boldsymbol{\tilde R}_{i,i} & \boldsymbol{\tilde R}_{i,i+1}\\  \boldsymbol{\tilde R}^\mathrm{T}_{i,i+1} &  \boldsymbol{ \tilde R}_{i+1,i+1}\end{array} \right]
 \in\R^{\tilde N_{2*i}\times\tilde N_{2*i}}.
 \end{equation*}
 Here, the sub-block $\boldsymbol{\tilde R}_{i_1, i_2}$ represents the correlation matrix of  $\boldsymbol{\tilde y}_{i_1}$ and $\boldsymbol{\tilde y}_{i_2}$, namely,
 \begin{equation*}
 \boldsymbol{\tilde R}_{i_1, i_2}= \left[\begin{array}{cc} \boldsymbol{R}_{i_1,i_2} & \partial\boldsymbol{R}_{i_1,i_2}\\  \partial\boldsymbol{R}^\mathrm{T}_{i_1,i_2} & \partial^2 \boldsymbol{ R}_{i_1,i_2}\end{array} \right]
\in\R^{\tilde N_{i_1}\times\tilde N_{i_2}},
\end{equation*}
 and its elements are given by
\begin{eqnarray*}
 \left(\boldsymbol{R}_{i_1, i_2}\right)_{k_1, k_2}&:=& R(\boldsymbol{x}^{(k_1)},{\boldsymbol{x}^{(k_2)}};\boldsymbol{\theta}),\\
\left(\partial\boldsymbol{R}_{i_1,i_2}\right)_{k_1, k_2}&:=&
\frac{\partial R(\boldsymbol{x}^{(k_1)},\boldsymbol{x}^{(k_2)};\boldsymbol{\theta})}{\partial {x^{(k_2)}_{l_1}}}, l_1=1,\ldots,n\\  \left(\partial^2\boldsymbol{R}_{i_1, i_2}\right)_{k_1, k_2}&: =&
 \frac{\partial^2 R(\boldsymbol{x}^{(k_1)},\boldsymbol{x}^{(k_2)};\boldsymbol{\theta})}{\partial {x^{(k_1)}_{l_1}}{\partial {x^{(k_2)}_{l_2}}}},l_1,l_2=1,\ldots,n. 
 \end{eqnarray*}
 where $k_1=\sum_{j=1}^{i_1-1} N_j+1,\ldots,\sum_{j=1}^{i_1} N_j$, $k_2=\sum_{j=1}^{i_2-1} N_j+1,\ldots,\sum_{j=1}^{i_2}N_j$.
 
  Taking the logarithm of $\hat f(\boldsymbol{y})$ from Eq. \eqref{2appendant}, we have 
  \begin{equation}
  \mathrm {ln} \hat f(\boldsymbol{y})=\sum_{i=1}^{m-1} \mathrm {ln}f(\boldsymbol{\tilde y}_{i},\boldsymbol{\tilde y}_{i+1})-\sum_{i=2}^{m-1} \mathrm {ln}f(\boldsymbol{\tilde y}_{i}).
  \label{sgelikelihood}
  \end{equation}
  
  Considering the hyper-parameters $\boldsymbol{\theta}$ as fixed and setting the partial derivatives of $\mathrm {ln} \hat f(\boldsymbol{y})$ with respect to $\tilde\beta_0$ and  $\tilde\sigma^2$ to zero, we obtain their optimal values analytically as
 \begin{eqnarray*}
  \tilde\beta_0 &=&\left(\sum_{i=1}^{m-1} \boldsymbol{\bar F}_{2*i}^{\rm T}\boldsymbol{\bar R}_{2*i}^{-1}\boldsymbol{\bar F}_{2*i}-\sum_{i=2}^{m-1} \boldsymbol{\bar F}_{i}^{\rm T}\boldsymbol{\bar R}_{i,i}^{-1}\boldsymbol{\bar F}_{i}\right)^{-1}  \\&&\hspace{4.5cm} \left(\sum_{i=1}^{m-1} \boldsymbol{\bar F}_{2*i}^{\rm T}\boldsymbol{\bar R}_{2*i}^{-1}\boldsymbol{\bar y}_{2*i}-\sum_{i=2}^{m-1} \boldsymbol{\bar F}_{i}^{\rm T}\boldsymbol{\bar R}_{i,i}^{-1}\boldsymbol{\bar y}_{i}\right), \\
 \tilde\sigma^2 &=& \frac{1}{(n+1)N}\Bigg(\sum_{i=1}^{m-1}(\boldsymbol{\tilde y}_{2*i}-\tilde\beta_0\boldsymbol{\tilde F}_{2*i})^{\rm T}\boldsymbol{\tilde R}_{2*i}^{-1}(\boldsymbol{\tilde y}_{2*i}-\tilde\beta_0\boldsymbol{\tilde F}_{2*i})-\\&&\hspace{4.9cm} \sum_{i=2}^{m-1}(\boldsymbol{\tilde y}_{i}-\tilde\beta_0\boldsymbol{\tilde F}_{i})^{\rm T}\boldsymbol{\tilde R}_{i,i}^{-1}(\boldsymbol{\tilde y}_{i}-\tilde\beta_0\boldsymbol{\tilde F}_{i})\Bigg).
\end{eqnarray*}

 Note that the optimal values of $\tilde\beta_0$ and $\tilde\sigma^2$ still depend on the hyper-parameters $\boldsymbol{\theta}$.
 Inserting their optimal values into Eq. (\ref{sgelikelihood}), we arrive at minimizing the  following likelihood function 
\begin{equation} \hat\ell(\boldsymbol{\theta})=(n+1)N\ln\tilde\sigma^2(\boldsymbol{\theta})+\sum_{i=1}^{m-1}\ln\left[{\rm det}\boldsymbol{\tilde R}_{2*i}(\boldsymbol{\theta})\right]-\sum_{i=2}^{m-1}\ln\left[{\rm det}\boldsymbol{\tilde R}_{i,i}(\boldsymbol{\theta})\right].
\label{likelihood}
\end{equation}
   
 As with standard GE-Kriging, one can modify the SGE-Kriging to become a regression model rather than an interpolator by adding the noise variance $\lambda$ to the diagonal elements of every small correlation matrix in Eq. \eqref{likelihood}, and the noise variance can be learned together with $\boldsymbol{\theta}$ by minimizing the likelihood in Eq. \eqref{likelihood}. 
     
\begin{figure}[htp]
\centering
\includegraphics[width=6 cm]{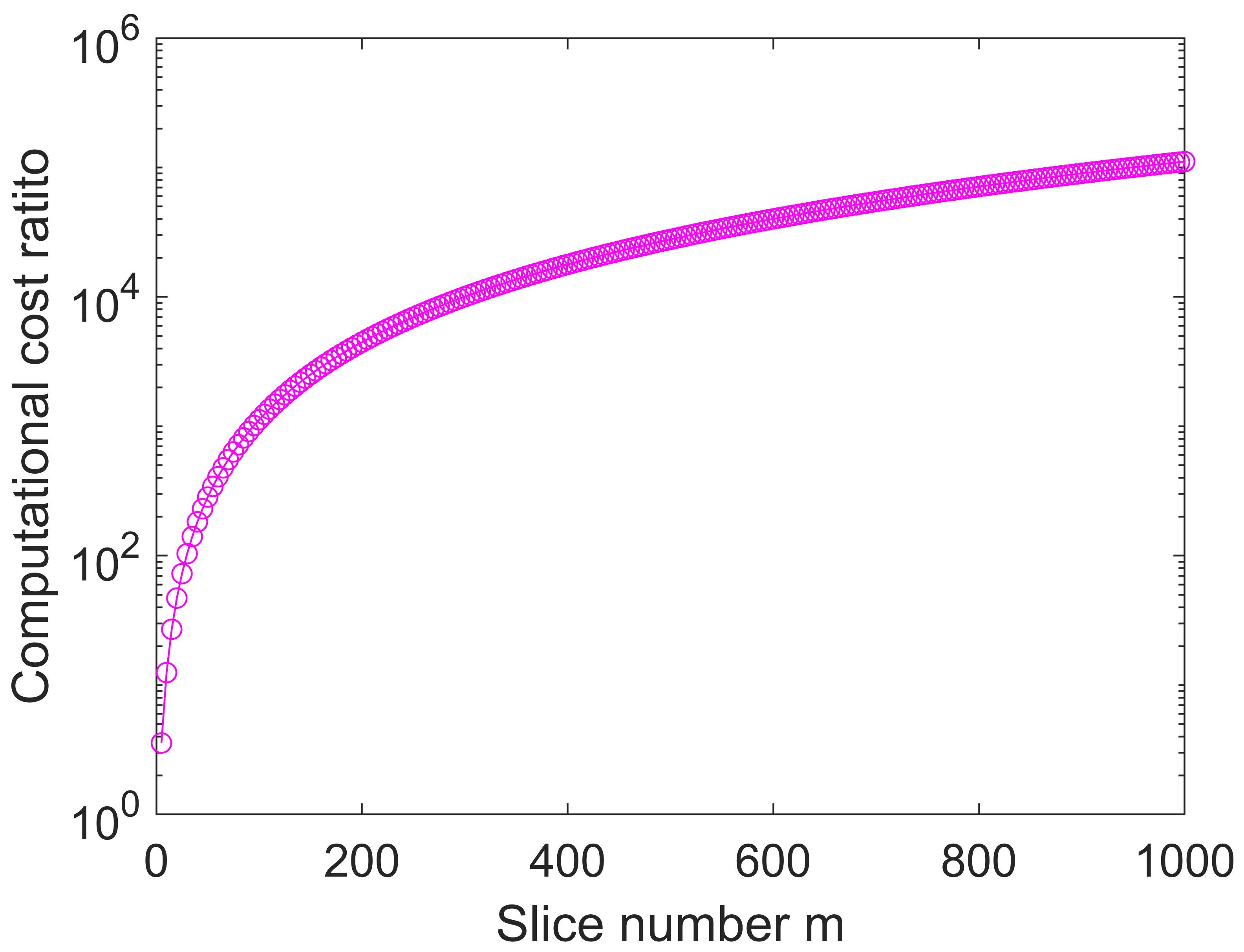}
\caption{Computational cost ratios for computing the Cholesky decomposition of the correlation matrices in GE-Kriging and SGE-Kriging.}
\label{slice_cost}
\end{figure}

  To evaluate this approximant of the likelihood function, the associated computational costs for decomposing $\boldsymbol{\tilde R}_{2*i}(i=1,\ldots,m-1)$ and $\boldsymbol{\tilde R}_{i,i}(i=2,\ldots,m-1)$ using the Cholesky method are $\mathcal{O}((m-1)(2n+2)^3(N/m)^3+(m-2)(n+1)^3(N/m)^3)$.
  We emphasize that the costs scale with the cubic term $\mathcal{O}((N/m)^3)$
  whereas the costs of evaluating the original likelihood function from Section \ref{sec:MLE} feature a cubic factor of  $\mathcal{O}(N^3)$.
  Therefore, the computational cost ratio for decomposing the correlation matrices of GE-Kriging and SGE-Kriging using the Cholesky method is given by
  \begin{equation}
    \frac{1}{8(m-1)/m^3+(m-2)/m^3} = \frac{m^3}{9m-10}
 \label{ratio}
 \end{equation}and therefore grows quadratically in $m$.
  In Table \ref{Table 1}, the computational cost ratio in Eq. \eqref{ratio} are listed for different values of $m$, and the curve of computational cost ratio with respect to $m$ is depicted in Fig. \ref{slice_cost}. It shows that the computational cost of SGE-Kriging is much lower than that of GE-Kriging for various cases, especially when the slice number $m$ is large. 
  
 \begin{table}[htbp]
 \centering
 \caption{Computational cost ratios for computing the Cholesky decomposition of the correlation matrices in GE-Kriging and SGE-Kriging.}
 \label{Table 1}
 \begin{tabular}{cc}
  \toprule  Slice number $m$ & Computational cost ratio \\
 \midrule
 10 & 12.50 \\
 20 & 47.06 \\  50 & 284.09 \\  100 & 1123.59 \\ 200 & 4469.27 \\
\bottomrule
\end{tabular}
\end{table}

  The difference between the original likelihood function in Eq. \eqref{reducedlikelihood} and the sliced likelihood function in Eq. \eqref{likelihood} is given by 
  \begin{equation}
  \begin{aligned}
  \ell(\boldsymbol{\theta})-\hat\ell(\boldsymbol{\theta}) =(n+1)N\ln\left(\frac{\sigma^2(\boldsymbol{\theta})}{\tilde\sigma^2(\boldsymbol{\theta})}\right)+\ln\left({\rm det}\boldsymbol{ R}(\boldsymbol{\theta})\frac{\prod_{i=2}^{m-1}{\rm det}\boldsymbol{\tilde R}_{i,i}(\boldsymbol{\theta})}{\prod_{i=1}^{m-1}{\rm det}\boldsymbol{\tilde R}_{2*i}(\boldsymbol{\theta})}\right).
  \label{difference}
  \end{aligned}
  \end{equation}

  From Eq. \eqref{difference}, one can see that the approximation error is determined by the process variance and the determinant of the correlation matrices, and they all depends on the hyper-parameters $\boldsymbol{\theta}$ and the slice number $m$. In theory, the difference tends to be small for problems with large $\boldsymbol{\theta}$ (highly nonlinear problems) and small slice number $m$, and vice versa. In practical application, one should set the slice number $m$ properly to balance the efficiency, accuracy and robustness. The performance of the approximated likelihood functions are investigated with two examples for different $\boldsymbol{\theta}$ and $m$ in Section \ref{sec:numex},. 
    
 \subsection{Cost-reduced hyper-parameter tuning}
 \label{sec:HPtuning}

 For high-dimensional problem\-s, it is still challenging to tune the hyper-parameters
 $\boldsymbol{\theta}=[\theta_1,\ldots,\theta_n]^{\rm T}$ of SGE-Kriging by minimizing the likelihood function in Eq. (\ref{likelihood}). In this subsection, we propose a way to utilize the global sensitivity analysis results to reduce the number of hyper-parameters $\boldsymbol{\theta}$.

 As mentioned earlier, the underlying  meaning of the hyper-parameters $\theta_k(k=1,\ldots,n)$ for a GE-Kriging model is related with the global sensitivity indices $S_k(k=1,\ldots,n)$ for a computational model $g(\boldsymbol{x})$. This observation suggests a way to determine the hyper-parameters by exploring their relationship with the global sensitivity indices. We note that a similar idea has been utilized in \cite{fu2020distance}, where the distance correlation \cite{da2015global} is used as global sensitivity measure, and the authors assumed a linear relationship between the aforementioned two quantities. However, working with a linear relationship may be too restrictive.
 Therefore, we propose to model the hyper-parameters with the following equation
 \begin{equation}
  \theta_k = h(\hat s_k) + \tilde\theta_k
  ,\quad k=1,\ldots,n, \label{hyper_model}
  \end{equation}
 where $ h(\hat s_k) = \alpha_1\hat s_k^{\alpha_2}+\alpha_3$ describes the trend of $\theta_k$. It provides a rough estimation of $\theta_k$ parameterized by three auxiliary parameters $\alpha_1$, $\alpha_2$ and $\alpha_3$, in which  $\hat s_k$ represents the normalized sensitivity index, namely, $\hat s_k= \hat S_k/\sum_{l=1}^n(\hat S_l)$. The correction term $\tilde\theta_k$ represents the residual. Note that in this way the trends of all hyper-parameters $\theta_k$ are determined by the same three parameters $\alpha_1,\alpha_2,\alpha_3$ and the associated sensitivity index. Obviously, this reparametrization of the model hyper-parameters only makes sense in cases where the parameter dimension is larger than three.

  In the current work, two approximate hyper-parameters optimization schemes are considered. 
  \textbf{Scheme 1} consists of two stages. First we optimize the trend parameterized by the $\alpha_i$ ($i=1,2,3$) globally, then we optimize the residuals locally.
  More precisely,  we firstly set the residuals $\tilde\theta_k(k=1,...,n)$ to 0, and the original high-dimensional likelihood function in Eq. \eqref{likelihood} can thus be replaced with the following 3-dimensional counterpart 
  \begin{equation}
  \ell(\boldsymbol{\alpha})=(n+1)N\ln\sigma^2(\boldsymbol{\alpha})+\sum_{i=1}^{m-1}\ln\left[{\rm det}\boldsymbol{\tilde R}_{2*i}(\boldsymbol{\alpha})\right]-\sum_{i=2}^{m-1}\ln\left[{\rm det}\boldsymbol{\tilde R}_{i,i}(\boldsymbol{\alpha})\right]
  \label{newlikelihood}
   \end{equation}
  where $\boldsymbol{\alpha}=[\alpha_1,\alpha_2,\alpha_3]^{\rm T}$. After the optimal $\boldsymbol{\hat\alpha}$ are found 
  with a global optimization method (in this work, we use the multi-starts gradient-free ``Hooke \& Jeeves'' pattern search method \cite{kowalik1968methods} for hyper-parameter tuning, which has been integrated into the well-known Kriging toolbox, DACE \cite{lophaven2002dace}), we turn to update the residuals $\tilde\theta_k(k=1,...,n)$. To this end, we set the initial values of $\theta_k$ as $\hat\alpha_1\hat s_k^{\hat\alpha_2}+\hat\alpha_3$, and determine the residuals by minimizing the high-dimensional likelihood function in Eq. \eqref{likelihood} with a local optimization method (single-start ``Hooke \& Jeeves'' algorithm). This step will calibrate the hyper-parameters locally and thus further improve the prediction accuracy of SGE-Kriging.
  
  For \textbf{Scheme 2},  we just omit the local residual optimization stage and keep the hyper-parameters $\hat\theta_k$ 
  obtained from the global optimizers $\hat \alpha_1, \hat\alpha_2,\hat\alpha_3$ of the 3-dimensional likelihood function in Eq.  \eqref{newlikelihood}.
  

  When the hyper-parameters $\boldsymbol{\hat\theta}$ of SGE-Kriging are determined, the final predictor of SGE-Kriging is consistent with the standard GE-Kriging in form, as presented in Eqs. (\ref{predictor})-(\ref{variancefunction}), in which $\beta_0$ and  $\sigma^2$ are given by Eqs. (\ref{beta}) and (\ref{sigma2}). However, one should note that we need to invert the correlation matrix $\boldsymbol{R}\in\R^{N(n+1)\times N(n+1)}$ with the Cholesky method one more time to eventually make predictions. 
 
 \section{Numerical examples}
  \label{sec:numex}
 In this section, we will demonstrate the effectiveness of SGE-Kriging on several benchmark problems, and its performance is compared to the  standard GE-Kriging and Kriging without incorporating derivative data. The relative mean squared error (RMSE) is used to measure the accuracy of various surrogate models, which is defined as
 \begin{equation}
 \label{eq:RMSE}
 {\rm RMSE} =  \frac{\sum_{i=1}^{N_1}\left(g(\boldsymbol {x}^{(i)})-\mu(\boldsymbol x^{(i)})\right)^2}{\sum_{i=1}^{N_1}\left(g(\boldsymbol {x}^{(i)})-\tilde g\right)^2},
 \end{equation}
 where $g(\boldsymbol {x}^{(i)})$ and $\mu(\boldsymbol x^{(i)})(i=1,\ldots,N_1)$ are the model response and the prediction of the surrogate model under consideration on the test sample points, respectively, and $\tilde g$ is the mean of the test sample response. In the experiments, we use $N_1=3000$.
    
As announced earlier, we scale the input space to a hypercube $[0,1]^n$, and the ``Hooke \& Jeeves'' algorithm with 10 random starts is used to optimize the hyper-parameters of the various surrogate models. For standard GE-Kriging and the Kriging model, the hyper-parameters $\theta_k(k=1,\ldots,n)$ are confined to the range of $[0.001,10]$. In SGE-Kriging, the auxiliary parameters $\alpha_1$, $\alpha_2$ and $\alpha_3$ are limited to intervals $[0.001,5]$, $[0.2,1]$ and $[0.001,5]$, respectively, so that the hyper-parameters $\theta_k (k=1,\ldots,n)$ of SGE-Kriging share the same bounds with those of GE-Kriging. Based on our own numerical experiments and the literature \cite{han2017weighted}, these settings support the robustness of the GE-Kriging and SGE-Kriging model. 
    
 \subsection{One-dimensional example}
 \label{sec:Ex1}
  In this subsection, the following
  one-dimensi\-onal analytical function is used to illustrate the effectiveness of SGE-Kriging 
  \begin{equation}
    g(\boldsymbol{x})=e^{-x}+{\rm sin}(5x)+{\rm cos}(5x)+0.2x+4, x\in [0,6].
  \end{equation}
 
  \begin{figure}[htp]
  \centering
  \vspace{0cm}
  \includegraphics[width=6cm]{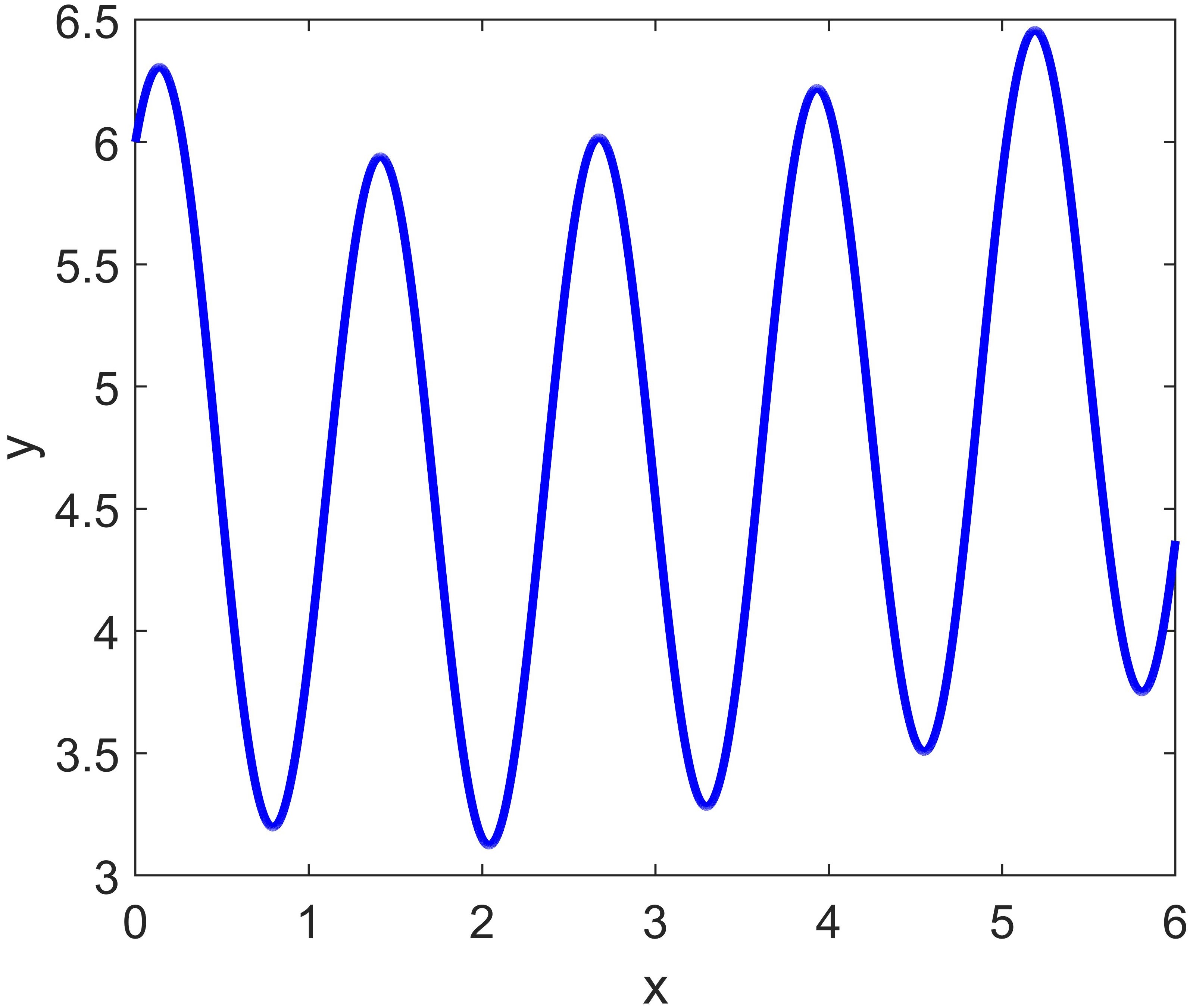}
  \caption{The true response in Example \ref{sec:Ex1}.}
  \label{example1_response}
  \end{figure}

\begin{figure}[htp]
  \centering
  \vspace{0cm}
  \includegraphics[width=12cm]{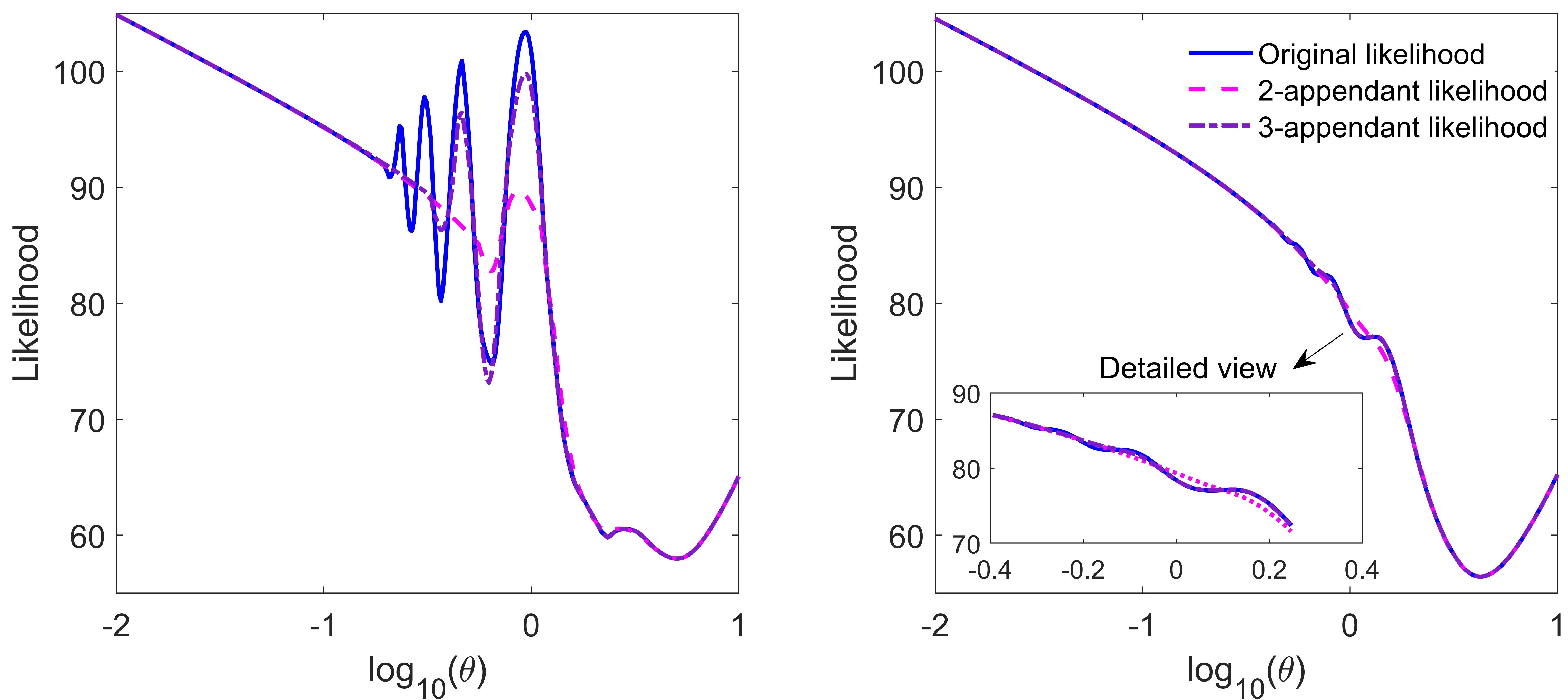}
  \caption{Comparison of likelihood functions of GE-Kriging and SGE-Kriging with $m=5$ using cubic spline correlation function (left) and biquadratic spline correlation function (right) for Example \ref{sec:Ex1}.}
  \label{example1_5slices}
  \end{figure}

\begin{figure}[htp]
  \centering
  \includegraphics[width=12cm]{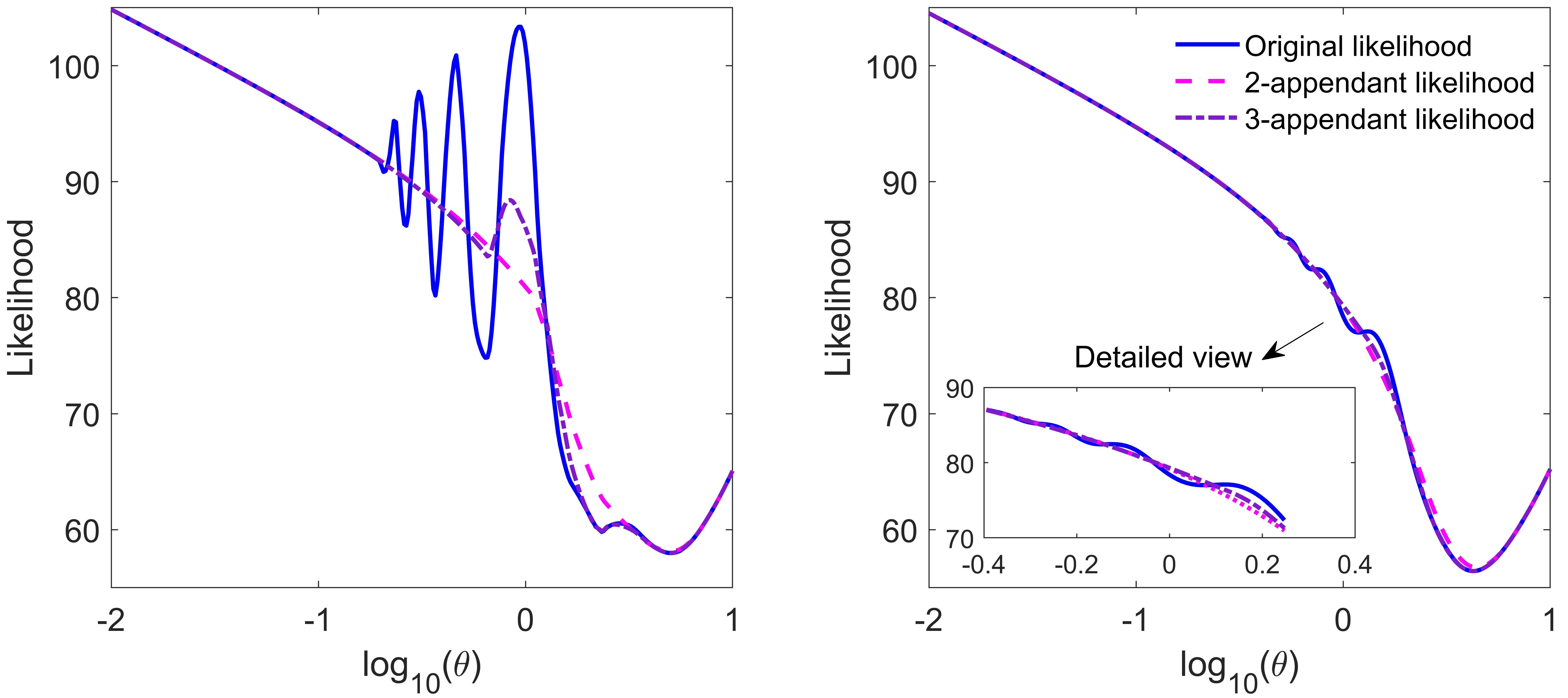}
   \caption{Comparison of likelihood functions of GE-Kriging and SGE-Kriging with $m=10$ for Example \ref{sec:Ex1} using the cubic spline correlation function (left) and the biquadratic spline correlation function (right).}
   \label{example1_10slices}
  \end{figure}

  The true response of the function is depicted in Fig. \ref{example1_response}. To fit it, we generate $10$ samples with the Latin Hypercube Sampling (LHS) method and construct GE-Kriging and SGE-Kriging predictors using the cubic and the biquadratic spline correlation functions. In SGE-Kriging, the sample set is divided into $m=5$ (two sample per slice) and $m=10$ (one samples per slice) slices equally, 
  and the sliced likelihood functions are depicted in Figs. \ref{example1_5slices} and \ref{example1_10slices}, respectively for comparison.
  
   For standard GE-Kriging, it is observed that the cubic spline function yields a likelihood function that exhibits a much more complex behavior with many local minima when compared to the biquadratic one, which suggests that the biquadratic spline function is more benign than the cubic spline one for applications in GE-Kriging. In contrast, the likelihood functions of SGE-Kriging exhibit a simpler shape when compared to the original one for both the two correlation functions with various settings, but they still feature the optimal value of the model hyper-parameter at virtually the same location. Additionally, as expected from the theory, the example shows that the likelihood functions of SGE-Kriging with less slices ($m=5$) or with considering the correlation of samples within more adjacent slices yield better approximations of the original likelihood function. Also, one can see that the 2-appendant likelihood function of SGE-Kriging with $m=10$ already provides satisfactory results, and the improvement of 3-appendant likelihood function is negligible when it comes to locating the global optimum in this example, especially when the biquadratic spline function is used. 
  
\begin{figure}[htp]
  \centering
  \vspace{0cm}
  \includegraphics[width=1.0\textwidth]{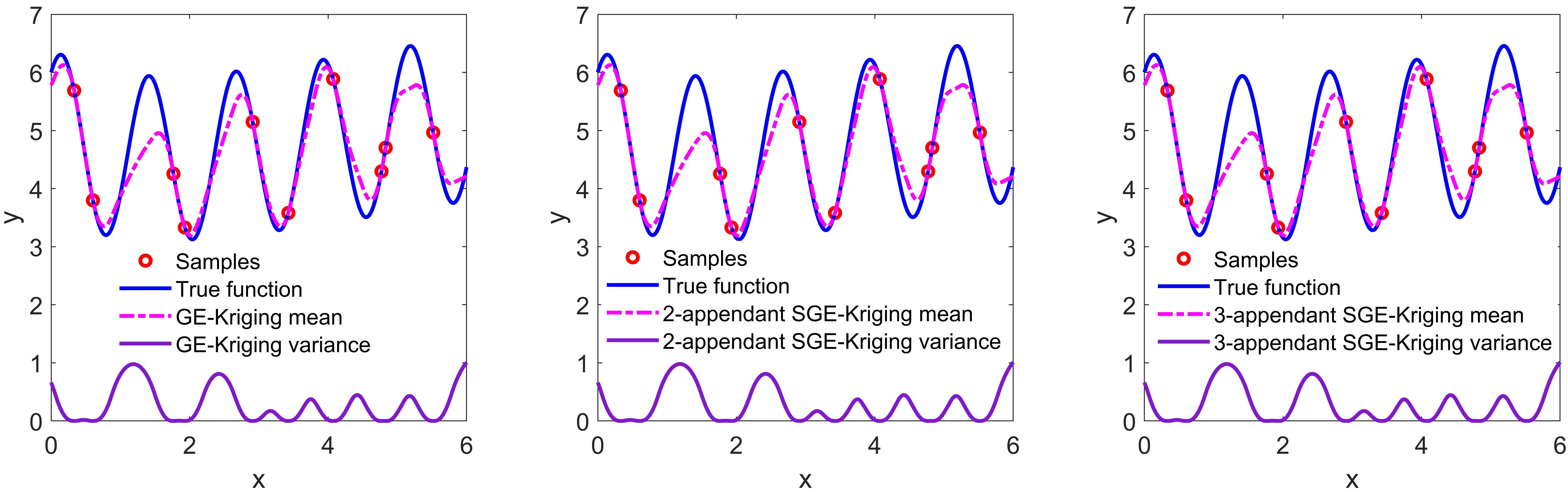}
  \caption{Comparisons of GE-Kriging and SGE-Kriging with $m=10$ using the cubic spline correlation function for Example \ref{sec:Ex1}}
  \label{example1_cubic10slices}
  \end{figure}

\begin{figure}[htp]
  \centering
  \vspace{0cm}
  \includegraphics[width=1.0\textwidth]{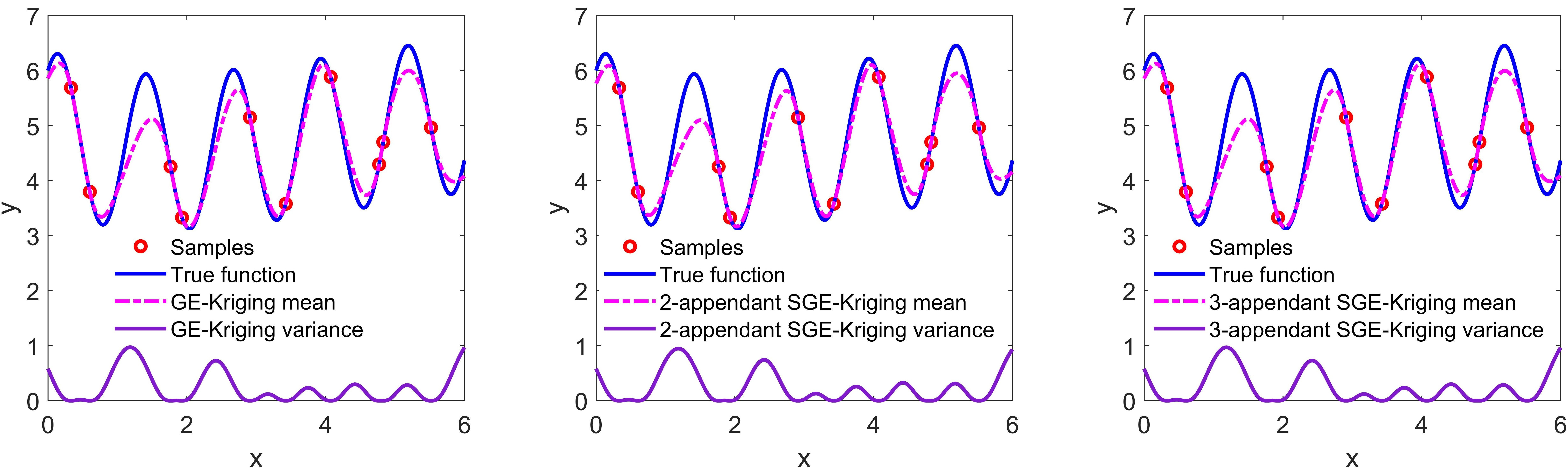}
  \caption{Comparison of GE-Kriging and SGE-Kriging with $m=10$ using biquadratic spline correlation function for Example \ref{sec:Ex1}}
   \label{example1_biquad10slices}
  \end{figure}

   The prediction accuracy of Kriging, GE-Kriging and SGE-Kriging with $m=10$ is depicted in Figs. \ref{example1_cubic10slices} and \ref{example1_biquad10slices} for comparison, and the detailed MSEs of the various surrogate models are listed in Table \ref{tab:mse_e1}. 
  Note that for this one-dimensional problem, the single hyper-parameter $\theta$ of SGE-Kriging is determined by minimizing the likelihood function in Eq. (\ref{likelihood}) directly.
   
   As expected, SGE-Kriging and GE-Kriging yield very similar predictions for both of the two correlation functions. Additionally, one can see that the GE-Kriging and SGE-Kriging predictors based on biquadratic spline correlation provide better results than the predictor that utilizes the cubic one, which supports that the biquadratic spline function is superior to the cubic one for applications in GE-Kriging and SGE-Kriging. 
  
 \begin{table}[htbp]
 \centering
\caption{Comparison of optimal hyper-parameters (obtained by full-factorial grid search)  and MSEs of GE-Kriging and SGE-Kriging with $m=10$ using different correlation functions for Example \ref{sec:Ex1}.
}
 \begin{tabular}{cccc}
 \toprule
  Surrogate model &  correlation function & Optimal $\theta$  & MSE \\
 \midrule
  GE-Kriging & cubic spline  & 3.7896 &  0.1263  \\ 
   GE-Kriging  & biquadratic spline & 3.9688 &  0.0875 \\ 
  2-appendant SGE-Kriging & cubic spline  & 3.7031 & 0.1272  \\
  2-appendant SGE-Kriging & biquadratic spline & 4.1565 & 0.0956  \\
  3-appendant SGE-Kriging & cubic spline & 3.7031 & 0.1272  \\
  3-appendant SGE-Kriging & biquadratic spline & 3.9688 & 0.0875 \\
  \bottomrule
\end{tabular}
 \label{tab:mse_e1}
 \end{table}
  
  \subsection{Two-dimensional example}
  \label{sec:2D_ex}
  In this subsection, a two-dimensional analytical function is used to demonstrate the effectiveness of the proposed SGE-Kriging model. The test function is defined as 
  \begin{equation}
    g(\boldsymbol{x})=(4-2.1x_1^2+\frac{x_1^4}{3})x_1^2+x_1x_2+(4x_2^2-4)x_2^2,
  \end{equation}
  where $x_1\in [-2,2]$ and $x_2\in [-1,1]$. 
  
\begin{figure}[htp]
\centering
\includegraphics[width=9cm]{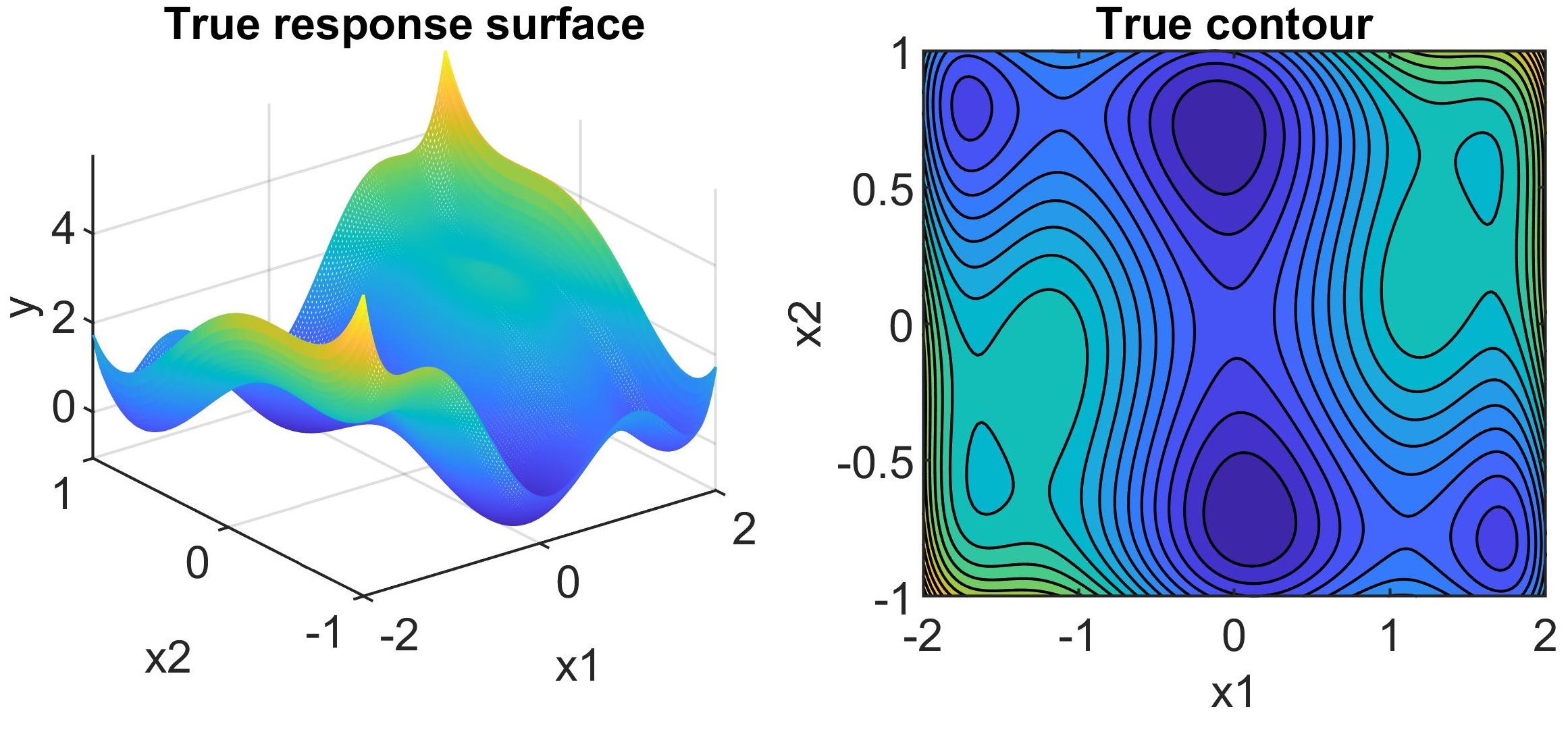}
\caption{ True response surface and contour of the function in Example \ref{sec:2D_ex}. }
\label{example2_response}
\end{figure}

   \begin{figure}[htp]
  \centering
  \includegraphics[width=1.0\textwidth]{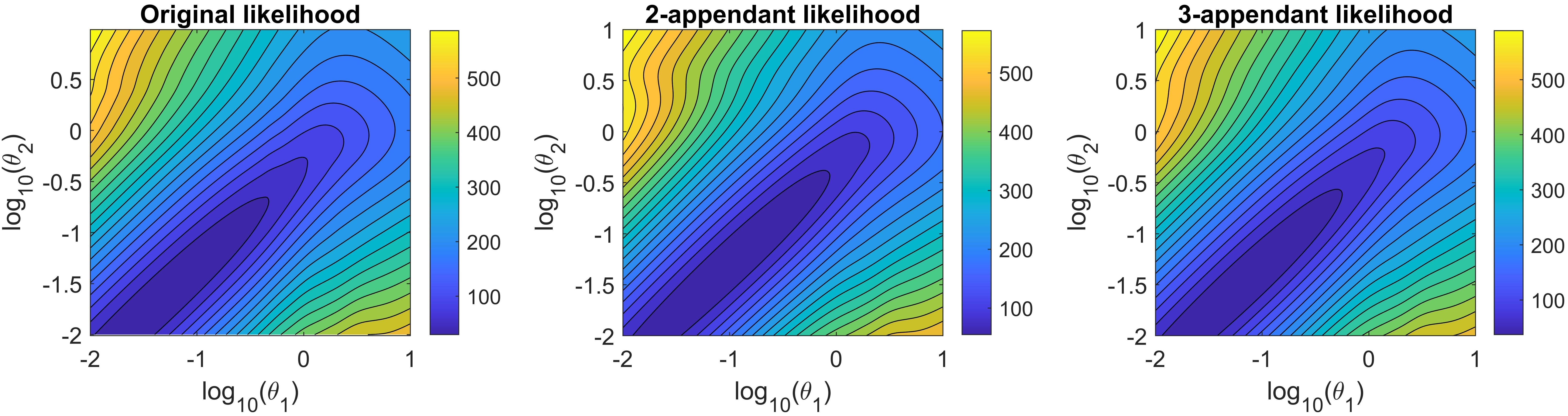}
  \caption{(cf. Example \ref{sec:2D_ex}) Comparison of the likelihood functions of GE-Kriging (left) and SGE-Kriging (middle, right) 
  based on the biquadratic spline correlation function with $m=5$.}
  \label{example2_5slices}
  \end{figure}
  
  \begin{figure}[htp]
  \centering
  \includegraphics[width=1.0\textwidth]{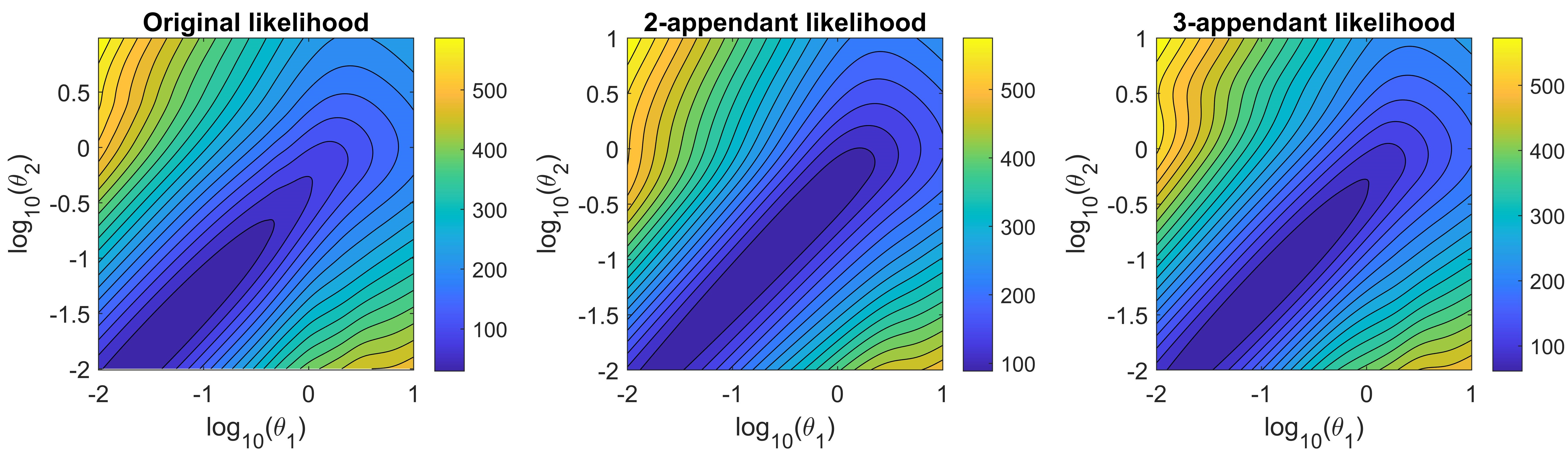}
  \caption{(cf. Example \ref{sec:2D_ex}) Same as Figure \ref{example2_5slices} but with SGE-Kriging with $m=10$.}
  \label{example2_10slices}
  \end{figure}
  
  \begin{figure}[htp]
  \centering
  \includegraphics[width=1.0\textwidth]{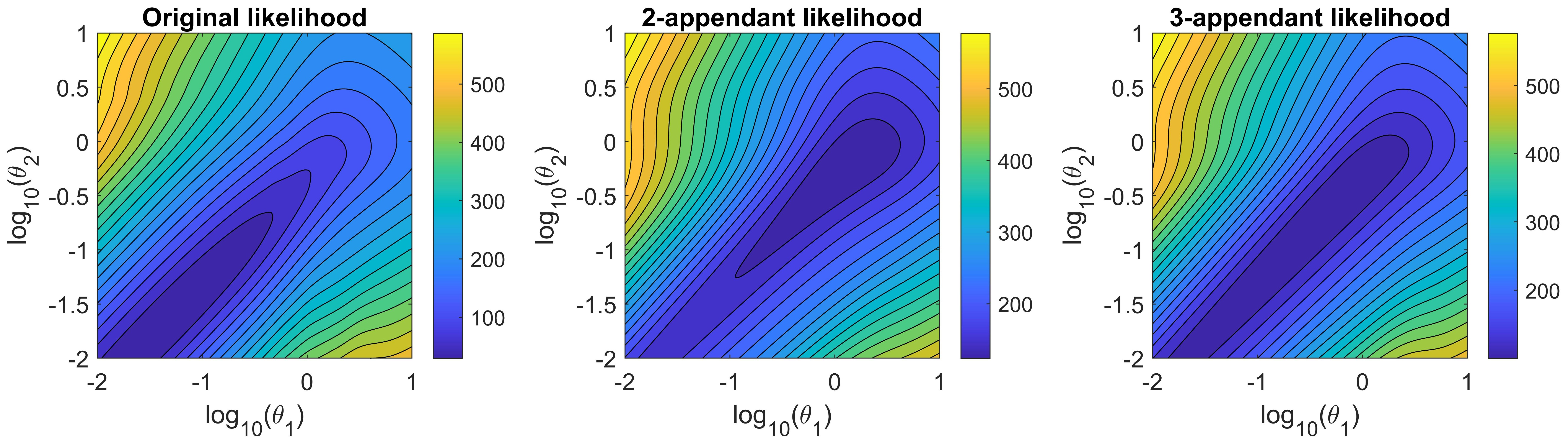}
  \caption{(cf. Example \ref{sec:2D_ex}) Same as Figure \ref{example2_5slices} but with SGE-Kriging with $m=20$.}
  \label{example2_20slices}
  \end{figure}
  
  \begin{table}[htbp]
  \centering
  \caption{Comparison of optimal hyper-parameters (obtained by grid search) and minimal likelihood function values of GE-Kriging and SGE-Kriging with different settings for Example 4.2.}
  \begin{tabular}{cccc}
  \toprule
   Likelihood function  &  $\theta_1$  & $\theta_2$ & Likelihood value \\
 \midrule
 Original likelihood &   0.0445 & 0.0222  &  28.9504\\ 2-appendant likelihood ($m=5$) &  0.0988 &  0.0477  & 42.1823 \\ 2-appendant likelihood ($m=10$) & 0.5053 &  0.2613 & 87.7040 \\  2-appendant likelihood ($m=20$) &  0.9437 &  0.4103  &  104.3997 \\
 3-appendant likelihood ($m=5$) & 0.0548 & 0.0273 & 31.4144 \\  3-appendant likelihood ($m=10$) &  0.1499 & 0.0749 & 58.9102\\ 3-appendant likelihood ($m=20$) & 0.6905 &  0.3449  &  97.1606\\ 
  \bottomrule
  \end{tabular}
  \label{tab:mse_e2}
  \end{table}

 The true response surface and its contour of the function in this subsection are depicted in Fig. \ref{example2_response}. To fit it, we generate $20$ samples with LHS randomly to train the various surrogate models. The likelihood functions of GE-Kriging and SGE-Kriging with different settings are depicted in Figs. \ref{example2_5slices}--\ref{example2_20slices}. It can be seen that the likelihood functions of SGE-Kriging mimics the global behavior of the original one quite well in all the considered cases. In Table \ref{tab:mse_e2}, the optimal hyper-parameters suggested by different likelihood functions and the corresponding minimal likelihood function values are listed. Again, we see that the likelihood functions of SGE-Kriging with less slices or with considering the correlation of samples within more adjacent slices yield better approximations of the original one. 


\begin{figure}[htp]
\centering
\includegraphics[width=9cm]{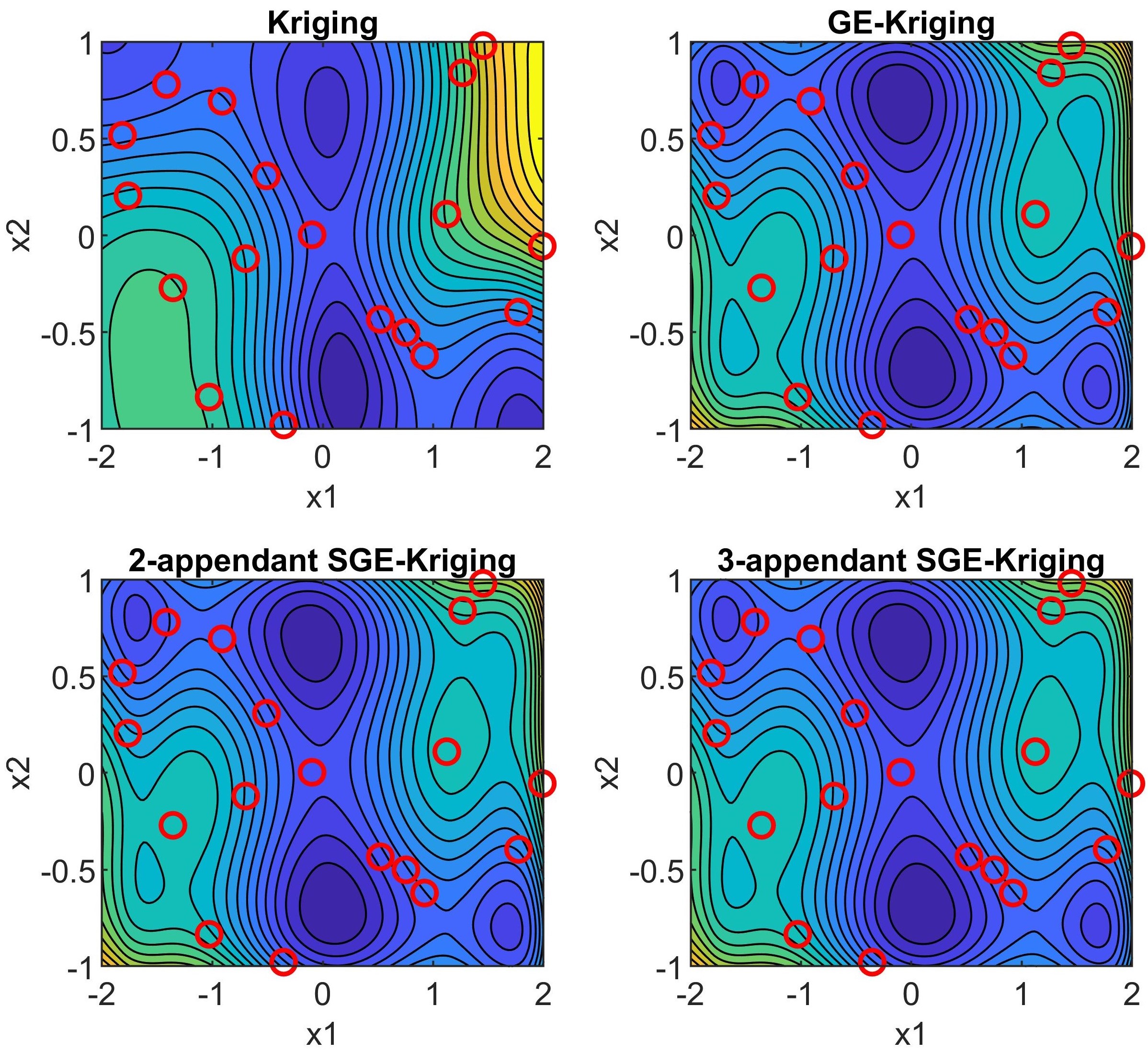}
\caption{(cf. Example \ref{sec:2D_ex})  Comparison of the prediction mean of Kriging, GE-Kriging and SGE-Kriging with $m=10$ using the biquadratic spline correlation function.}
\label{example2_mean}
\end{figure}

 \begin{figure}[htp]
\centering
\includegraphics[width=12cm]{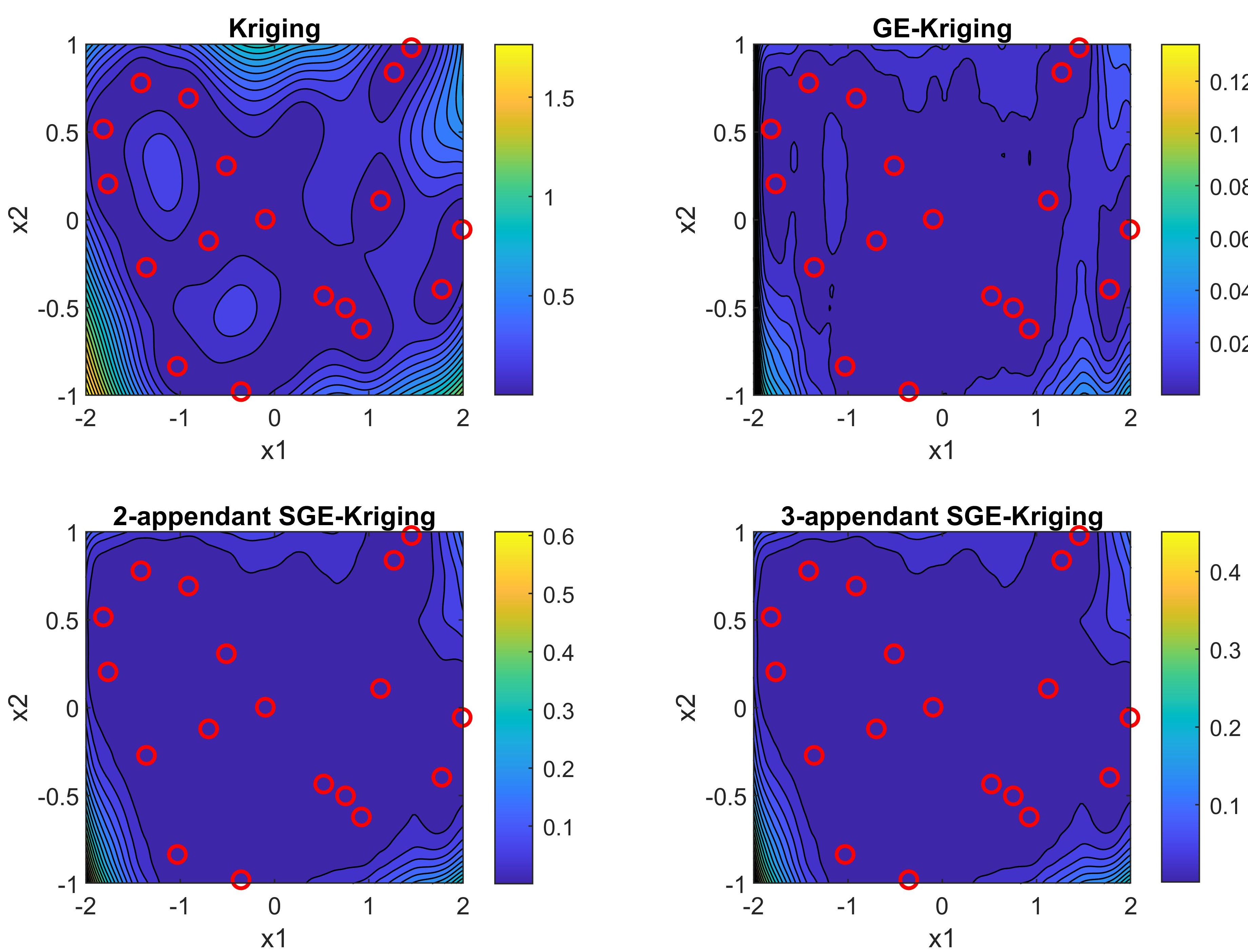}
\caption{(cf. Example \ref{sec:2D_ex}) Comparison of the prediction variance of Kriging, GE-Kriging and SGE-Kriging with $m=10$ using the biquadratic spline correlation function.}
\label{example2_variance}
\end{figure}

   The true response surface of this function, the corresponding contour and the prediction mean values of Kriging, GE-Kriging and SGE-Kriging with 10 total slices are depicted in Fig. \ref{example2_mean}.
   For this two-dimensional problem, the hyper-parameters of SGE-Kriging are still tuned by minimizing the likelihood function in Eq. (\ref{likelihood}). 
   It shows that SGE-Kriging and GE-Kriging provide similar predictions, and both of them outperform Kriging significantly. Specifially, the MSEs of Kriging, GE-Kriging, 2-appendant and 3-appendant SGE-Kriging are 0.1140, 0.00175, 0.00306 and 0.00212, respectively, which confirms the effectiveness of SGE-Kriging. Moreover, in Fig. \ref{example2_variance}, we see that the prediction variance of SGE-Kriging is slightly higher than that of GE-Kriging. Indeed, since part of the sample correlation is neglected, the accuracy of SGE-Kriging is expected to be inferior to GE-Kriging in theory. Note that for this low-dimensional case, the computational cost for training different surrogate models are negligible, and thus they are not provided for comparison.

 \subsection{8-dimensional Borehole function}
 \label{sec:Borehole}
  In this subsection, we demonstrate the performance of the hyper-parameter tuning equation in Eq. \eqref{hyper_model} on the Borehole function \cite{moon2012two}, which is defined as
  \begin{eqnarray*}
   g(\boldsymbol{x})=\frac{5T_u(H_u-H_l)}{\ln (r/r_w)(1.5+\frac{2LT_u}{\ln(r/r_w)r_w^2K_w})+\frac{T_u}{T_l}}, 
 \end{eqnarray*}
 where $\boldsymbol{x} = [r_w,r,T_u,H_u,H_l,L,K_w]^{\rm T}$, and the physical meaning as well as input range of input variables are listed in Table \ref{tab:Borehole_inputs}. 
 
 \begin{table}
 \caption{Physical meanings and ranges of the input variable in Borehole function}
 \begin{tabular}{ccc}
  \toprule
   Input variable  &  Physical meaning   & Input range  \\
 \midrule
 $r_w$ &  radius of borehole ($m$) & $[0.05,0.15]$ \\ $r$ &  radius of influence ($m$) &   $[100,50000]$  \\ 
 $T_u$ & transmissivity of upper aquifer ($m^2/yr$) &  $[63070,115600]$ \\  
 $H_u$ &  potentiometric head of upper aquifer ($m$) &  $[990,1110]$   \\
 $T_l$ & transmissivity of lower aquifer ($m^2/yr$) & $[63.1,116]$  \\ 
 $H_l$ &  potentiometric head of lower aquifer ($m$)  & $[700,820]$\\
 $L$ & length of borehole ($m$)   & $[1120,1680]$ \\ 
 $K_w$ & hydraulic conductivity of borehole ($m^2/yr$)   &  $[9855,12045]$\\ 
  \bottomrule
  \end{tabular}
  \label{tab:Borehole_inputs}
  \end{table}
  
\begin{figure}[htp]
\centering
\includegraphics[width=7.5 cm]{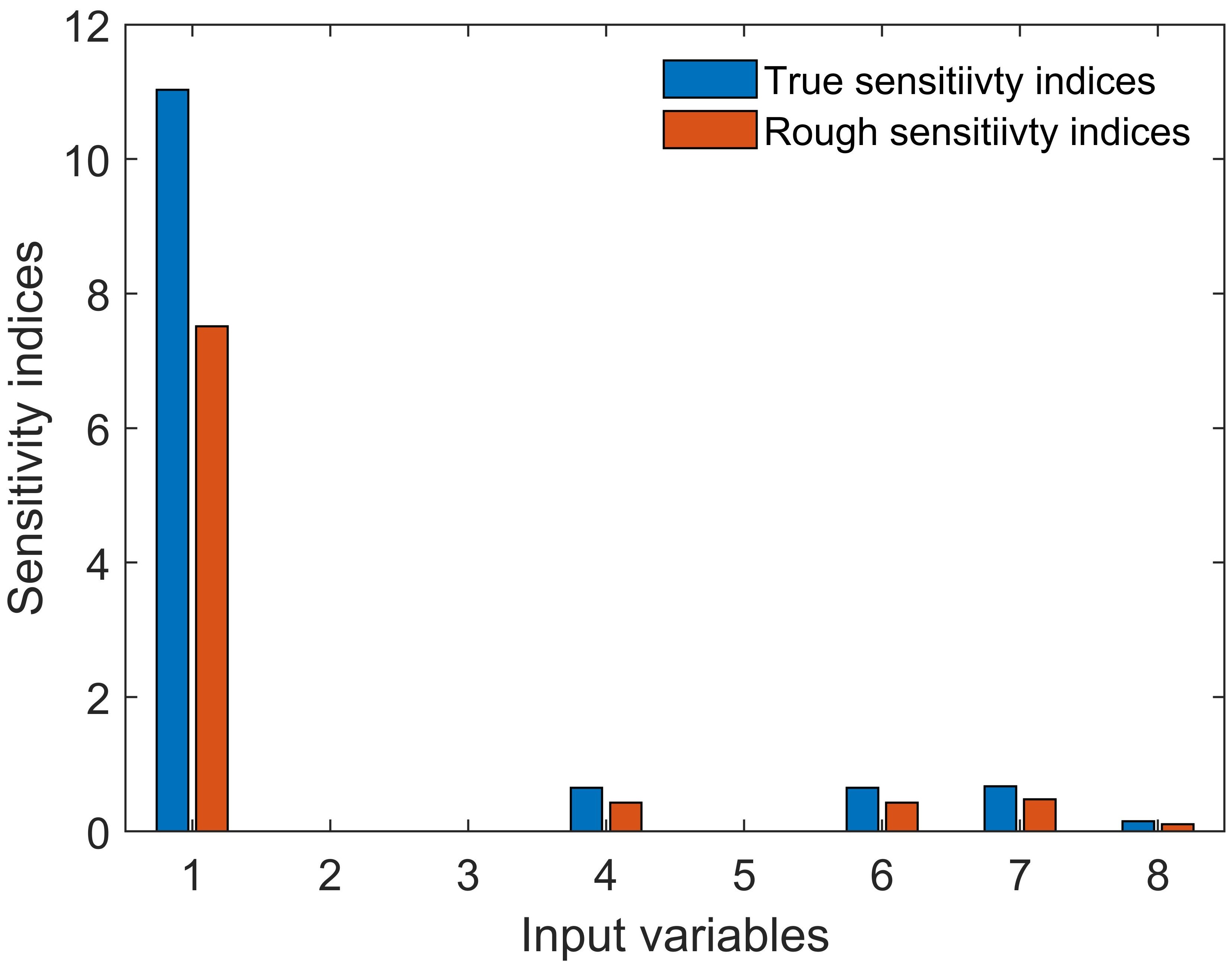}
\caption{Derivative-based global sensitivity indices of the Borehole function; Accurate sensitivity indices are obtained with 10000 samples, and the estimated ones are obtained with 20 samples. }
\label{Borehole_sensitivity}
\end{figure}

  In this example, we generate 20 samples to train various surrogate models for the Borehole function. To assess the 
  dependency of the parameterized Eq. \eqref{hyper_model} on the accuracy of the derivative-based global sensitivity indices, we optimize the hyper-parameters for various surrogate models with both the true sensitivity indices (as estimated with 10000 samples) and the rough sensitivity indices (as estimated with 20 training samples). The derivative-based global sensitivity indices are presented in Fig. \ref{Borehole_sensitivity}. One can see that the sensitivity indices estimated with 20 samples, while not being an accurate match for the true ones, still can rank the importance of the input variables relatively well.
  
\begin{figure}[htp]
 \centering
\includegraphics[width=8 cm] {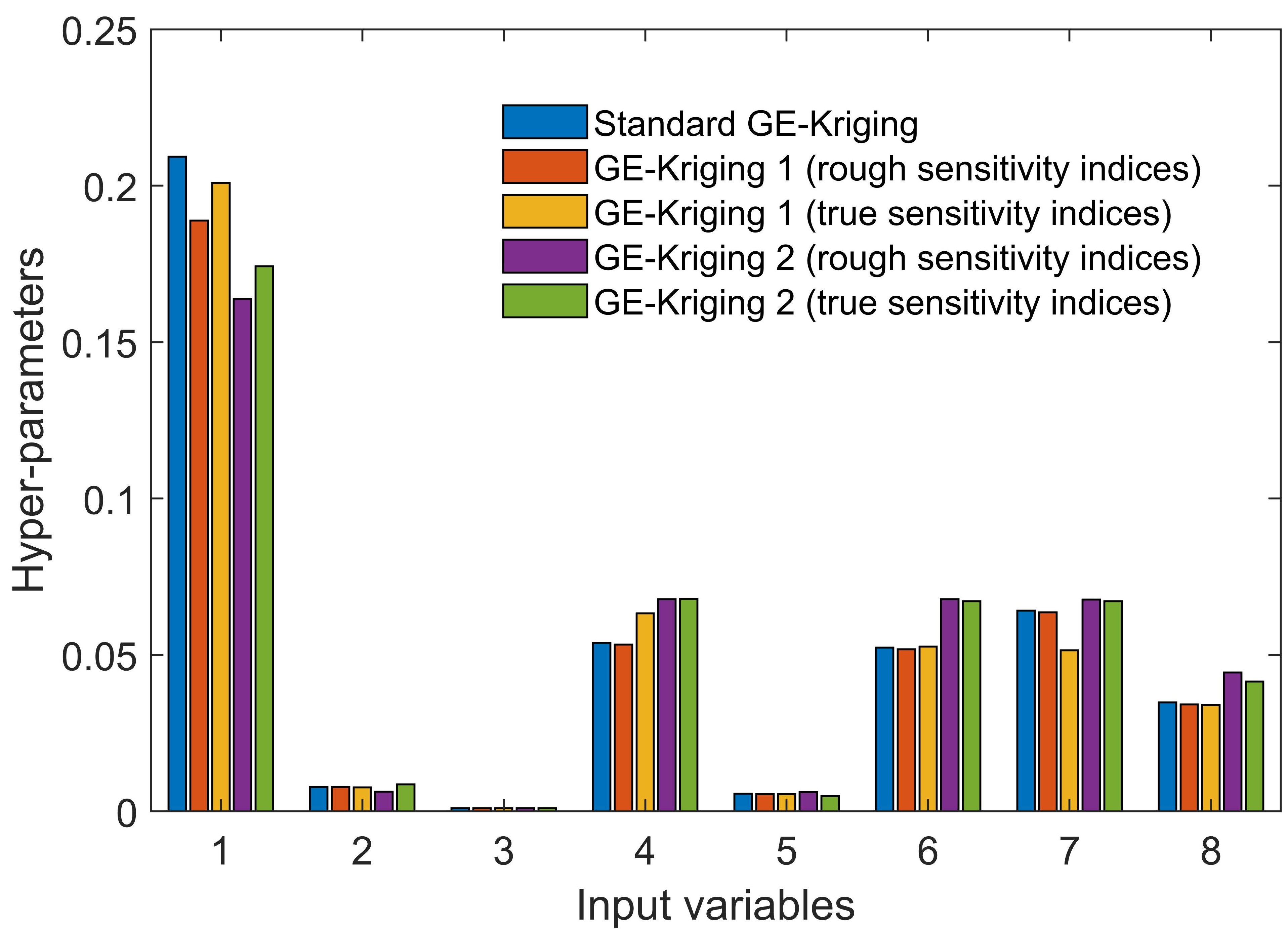}
\caption{Hyper-parameters of GE-Kriging models with different settings for the of Borehole function.}
\label{Borehole_hyperparameter_GEK}
\end{figure}

\begin{figure}[htp]
\centering
\includegraphics[width=8 cm]{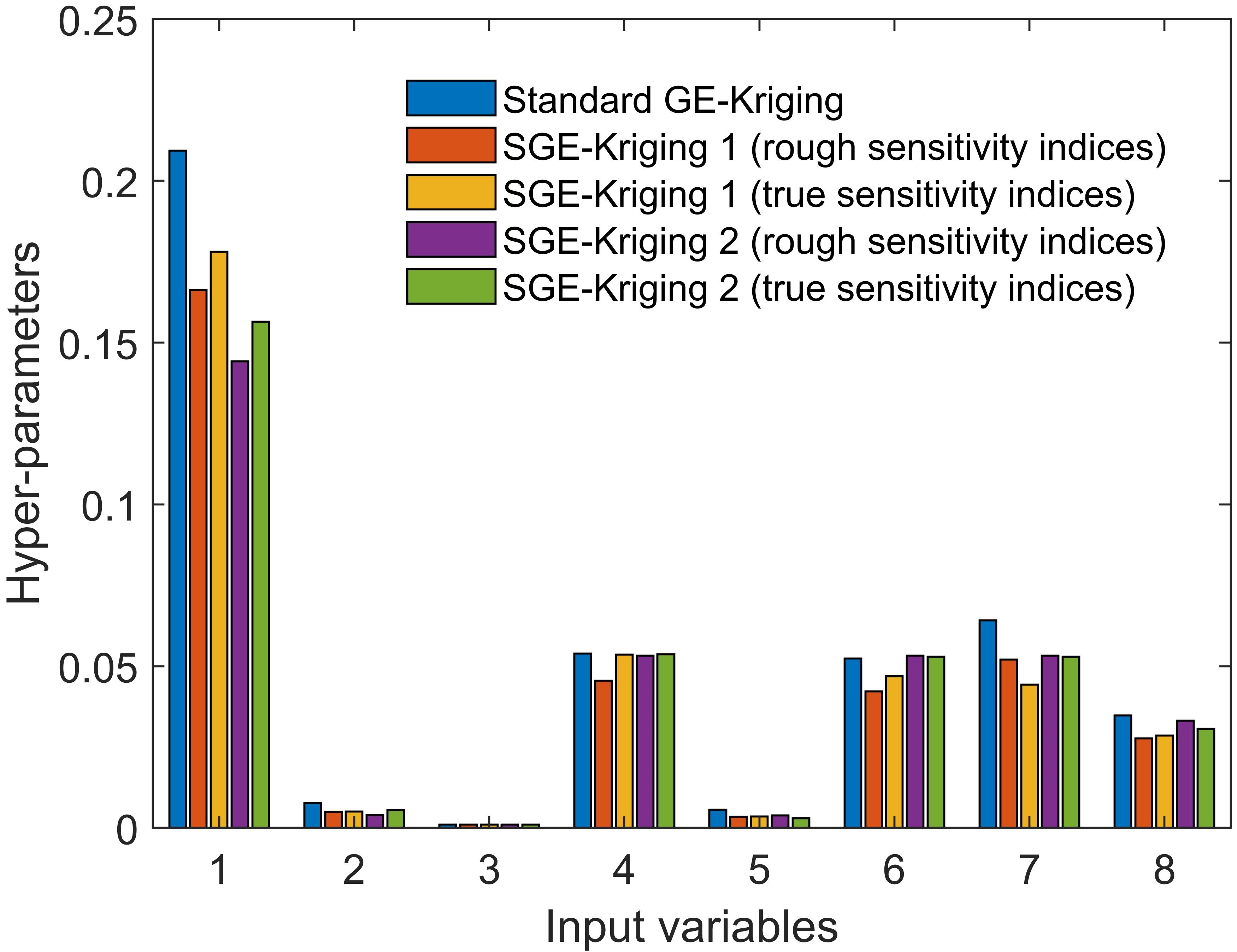}
\caption{Hyper-parameters of SGE-Kriging models with different settings for the of Borehole function.}
\label{Borehole_hyperparameter_SGEK}
\end{figure}

   We train GE-Kriging predictors and 2-appendant SGE-Kriging predictors ($m=10$) with the two hyper-parameter tuning schemes introduced in Section \ref{sec:HPtuning}. For comparison, we also 
  optimize the hyper-parameters $\boldsymbol{\theta}$ directly with a numerical optimization algorithm, in the following referred to as the 
  standard scheme.
  The detailed settings are listed in Table \ref{tab:Borehole_settings}. The hyper-parameters of the various GE-Kriging and SGE-Kriging models are depicted in Fig. \ref{Borehole_hyperparameter_GEK} and Fig. \ref{Borehole_hyperparameter_SGEK}, respectively. The precise numerical results for the models under consideration are listed in Table \ref{tab:Borehole_comparison}.
  It can be seen that the proposed 3-parameter equation \eqref{hyper_model} is effective for tuning the 8 hyper-parameters of GE-Kriging and SGE-Kriging. This holds both for the accurate sensitivity indices and the rough ones. The hyper-parameters obtained with the true sensitivity indices are slightly closer to that obtained with standard GE-Kriging, and thus the corresponding likelihood better matches the one of Standard GE-Kriging. For the Borehole function, one can see that the hyper-parameter tuning scheme 2, namely, the one without updating the residuals, already provides acceptable results. The accuracy of both GE-Kriging and SGE-Kriging improves only slightly by updating the residuals locally. In addition, it shows that the sliced likelihood values ($m=10$) in SGE-Kriging are larger than the true likelihood values in GE-Kriging. 
  (Recall, that maximizing the likelihood was reformulated as a minimization problem, so that ``larger'' means ``worse''.) However, the performance of SGE-Kriging and GE-Kriging is comparable. This example demonstrates the effectiveness of both the sliced likelihood function and the 3-parameters equation for hyper-parameter tuning proposed in Section \ref{sec:HPtuning}.  
 \begin{table}[htbp]
  \centering
  \caption{Settings for tuning the different surrogate model for the Borehole function:
  `Standard scheme' means that a numerical optimizer for the $\theta$-hyper paramters is employed. `Scheme 1' works with the three-dimensional reparameterization of \eqref{hyper_model} plus a local tuning of the associated residuals. ´Scheme 2' starts as ´Scheme 1' but omits the local
  residual optimization.}
  \begin{tabular}{ccc}
  \toprule
  Model & Likelihood & Hyper-parameter tuning scheme  \\
 \midrule
  Standard Kriging  &  True likelihood & Standard scheme \\  Standard GE-Kriging &  True likelihood & Standard scheme \\ GE-Kriging 1 & True likelihood  & Scheme 1 in Section \ref{sec:HPtuning} \\  GE-Kriging 2 &  True likelihood   &  Scheme 2  in Section \ref{sec:HPtuning} \\  SGE-Kriging 1  &  Sliced likelihood  &  Scheme 1 in Section \ref{sec:HPtuning} \\  SGE-Kriging 2  &    Sliced likelihood  & Scheme 2  in Section \ref{sec:HPtuning}\\
  \bottomrule
  \end{tabular}
  \label{tab:Borehole_settings}
  \end{table}
  
  \begin{table}[htbp]
  \centering
  \caption{Comparison of various surrogate models with different settings using 20 training samples for the Borehole function}
  \begin{tabular}{cccc}
  \toprule
  Model & RMSE  & $(\alpha_1,\alpha_2,\alpha_3)$ & likelihood \\
 \midrule
  Standard Kriging  &  0.0302 & -  & -59.372\\  Standard GE-Kriging &  $5.939\times10^{-4}$ & - &   -$1.205\times10^{3}$ \\ GE-Kriging 1 \\ (rough sensitivity indices) & $6.016\times10^{-4}$  & (0.1716,0.3110,0.0010) &   -$1.204\times10^{3}$ \\  GE-Kriging 1 \\ (true sensitivity indices) &   $5.934\times10^{-4}$  & (0.1839,0.3398,0.0010) &   -$1.205\times10^{3}$ \\  GE-Kriging 2 \\ (rough sensitivity indices) &   $6.717\times10^{-4}$  & (0.1716,0.3110,0.0010)  &   -$1.183\times10^{3}$ \\ GE-Kriging 2 \\ (true sensitivity indices) &   $6.413\times10^{-4}$  & (0.1839,0.3398,0.0010) &   -$1.190\times10^{3}$ \\  SGE-Kriging 1 \\ (rough sensitivity indices) & $6.099\times10^{-4}$  & (0.1520,0.3516,0.0010) &   -$1.123\times10^{3}$ \\  SGE-Kriging 1 \\ (true sensitivity indices) &   $5.967\times10^{-4}$  & (0.1664,0.3869,0.0010) &   -$1.123\times10^{3}$ \\  SGE-Kriging 2 \\ (rough sensitivity indices) &   $6.409\times10^{-4}$  & (0.1520,0.3516,0.0010)  &   -$1.109\times10^{3}$ \\  SGE-Kriging 2 \\ (true sensitivity indices) &   $6.074\times10^{-4}$  & (0.1664,0.3869,0.0010) &   -$1.116\times10^{3}$ \\  
  \bottomrule
  \end{tabular}
  \label{tab:Borehole_comparison}
  \end{table}

 \subsection{30-dimensional Rosenbrock function}
\label{sec:Rosenbrock}
 In this subsection, we demonstrate the performance of SGE-Kriging on the Rosenbrock function \cite{han2017weighted}{}, which is defined as
\begin{eqnarray*} g(\boldsymbol{x})=\sum_{i=1}^{n-1}(x_i-1)^4+\sum_{i=2}^n\sqrt{i}(x_i-x_{i-1}^2)^2, x_i\in[-1,1],(i=1,\ldots,n)
 \end{eqnarray*}

\begin{figure}[htp]
\centering
\includegraphics[width=13 cm]{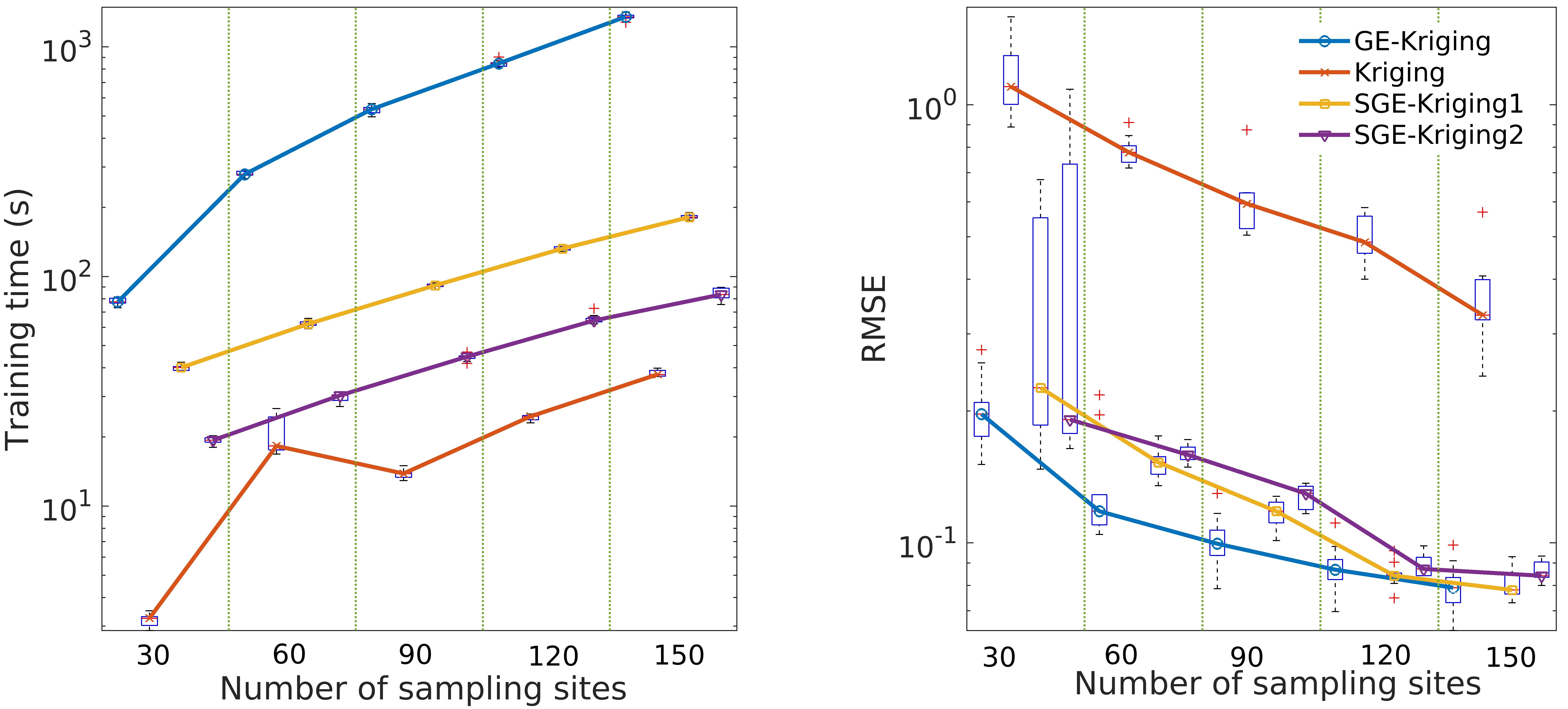}
\caption{Comparison of the performance of Kriging, GE-Kriging and SGE-Kriging (SGE-Kriging1 = updating the residual terms, SGE-Kriging2 = neglecting the residual terms) with $m=10$ using the biquadratic spline correlation function for a 30-dimensional Rosenbrock test function. Left: training time versus number of sample sites. Right: RMSE according to Eq. \eqref{eq:RMSE} versus number of sample sites.}
\label{Rosenbrock}
\end{figure}

 We set the input dimension and the slice number to  $n=30$ and $m=10$ respectively. The RMSEs and training times of GE-Kriging, SGE-Kriging and Kriging versus the number of sample sites are depicted in Fig. \ref{Rosenbrock}.
 Note that all the calculations are repeated $10$ times to level out the effects of randomness in the training sample set, and all the numerical experiments are performed on the same HPC-cluster on the UCloud platform (Inter Xeon Gold 6130 CPU, 64-cores, 2.1GHz)
 \tc{managed by the Danish e-infrastructure cooperation DeiC.}
    
  The box plots in Fig. \ref{Rosenbrock} are provided to illustrate the robustness of the different methods. In these plots, the central mark of each box indicates the median, and the bottom and top edges of the box indicate the $25$th and $75$th percentiles, and  outliers are plotted individually using the '+' marker. The plots show that the accuracy of SGE-Kriging is comparable to standard GE-Kriging. In addition, one can see that the two hyper-parameter tuning methods from Section \ref{sec:HPtuning} yield similar results.
  
 
  This example demonstrates that the  hyper-parameter $\theta_k$ are well described by the sensitivity information and the three-parameters equation $\alpha_1\hat s_k^{\alpha_2}+\alpha_3$. The contribution of the residual terms is marginal. One can see that to train a GE-Kriging model is time-consuming: It takes more than $1355$s to train a GE-Kriging model for a $30$-dimensional Rosenbrock function with 150 sample sites.
  Compared to GE-Kriging, the training time of SGE-Kriging is reduced sharply. Specifically, using $150$ sample points, training a SGE-Kriging 1 and SGE-Kriging 2  are $7.5$ and  $16.2$ times faster than GE-Kriging respectively. 
  
  Yet, the example also shows that SGE-Kriging exhibits a bad approximation quality, when the total number of sample sites is small. In these cases, the sample sites within each slice is too small, and the sliced likelihood function cannot approximate the original one very well. However, the performance improves considerably with increasing the number of sample sites.
  
 \subsection{High-dimensional Dixon-Price function}
  \label{sec:Dixon-Price}
  
   In this subsection, a high-dimensional Dixon-Price  function \cite{zhao2020efficient} is employed to further demonstrate the performance of SGE-Kriging. It is defined as
  \begin{eqnarray}
  g(\boldsymbol{x})=(x_1-1)^2+\sum_{i=2}^ni(2x_i^2-x_{i-1})^2, x_i\in[0,1],(i=1,\ldots,n). 
   \end{eqnarray} 
  
 \begin{figure}[htp]
\centering
\includegraphics[width=13 cm]{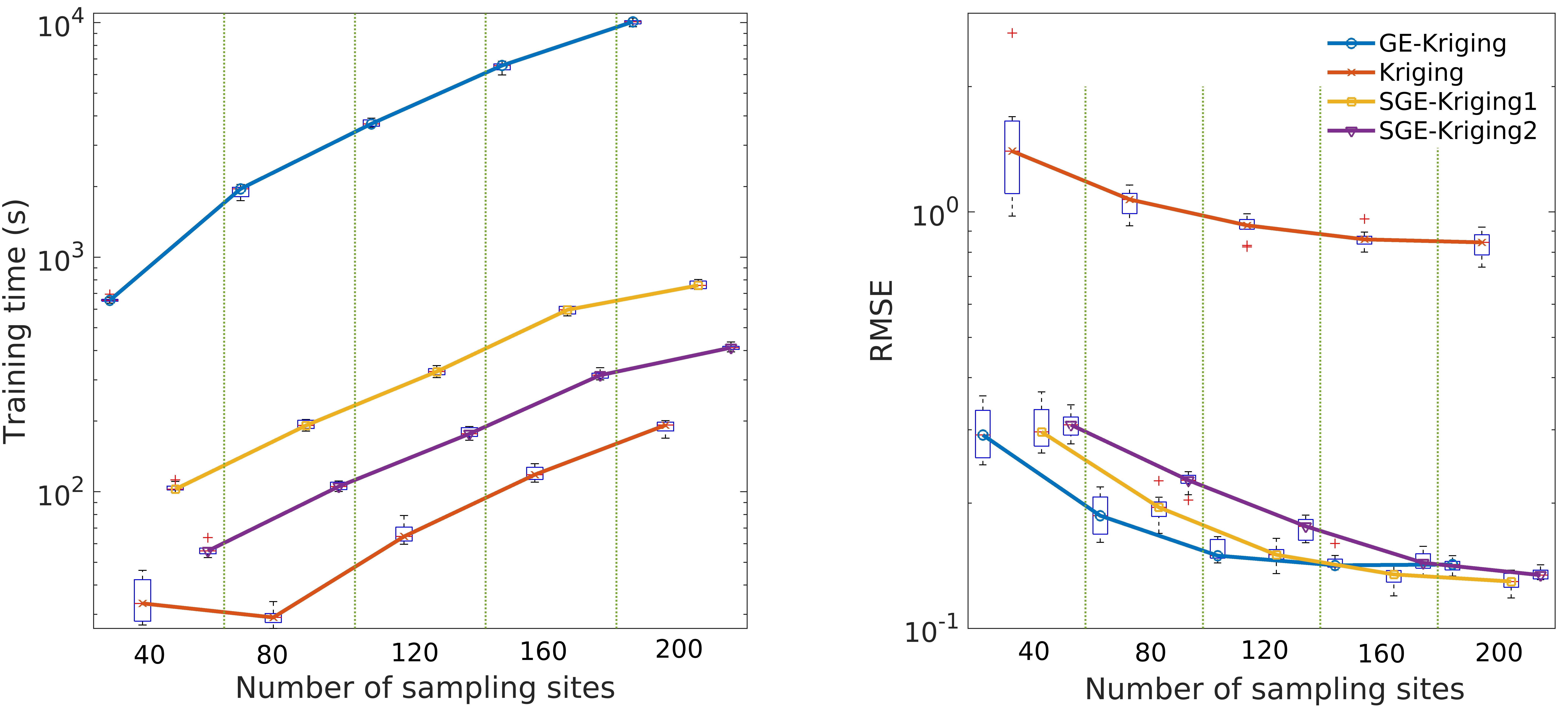}
\caption{Comparison of the performance of Kriging, GE-Kriging and SGE-Kriging (SGE-Kriging1 =  updating the residual terms, SGE-Kriging2 = neglecting the residual terms) with $m=10$ using the biquadratic spline correlation function for a 50-dimensional Dixon-Price test function. Left: training time versus number of sample sites. Right: RMSE according to Eq. \eqref{eq:RMSE} versus number of sample site.}
\label{Dixon-Price}
\end{figure}
  
In this example, the input dimension is set as $n=50$, and we train various surrogate models with the same settings as in Example \ref{sec:Rosenbrock}. 
The RMSEs and training times of GE-Kriging, SGE-Kriging and Kriging versus the number of sample sites are depicted in Fig. \ref{Dixon-Price}. It shows that SGE-Kriging provides a similar predictor compared to the standard GE-Kriging model, and both of them clearly outperform the Kriging model. Again, we see that the two hyper-parameters tuning schemes introduced in Section \ref{sec:HPtuning} yield very similar SGE-Kriging predictors, and the contribution of the residuals is negligible. 
 Regarding the timings, it takes about $1\times10^4$s ($\approx 2.8h$) to train a GE-Kriging model for a $50$-dimensional Dixon-price function with $150$ sample sites. In contrast, training a SGE-Kriging1 predictor with the same settings is about $25$ times faster than training GE-Kriging. This example demonstrates again that SGE-Kriging provides a satisfactory balance between accuracy and efficiency. 
  
\subsection{ Aerodynamic Modeling of NACA 0012 airfoil}
\begin{figure}[htp]
\centering
\includegraphics[width=0.8\textwidth]{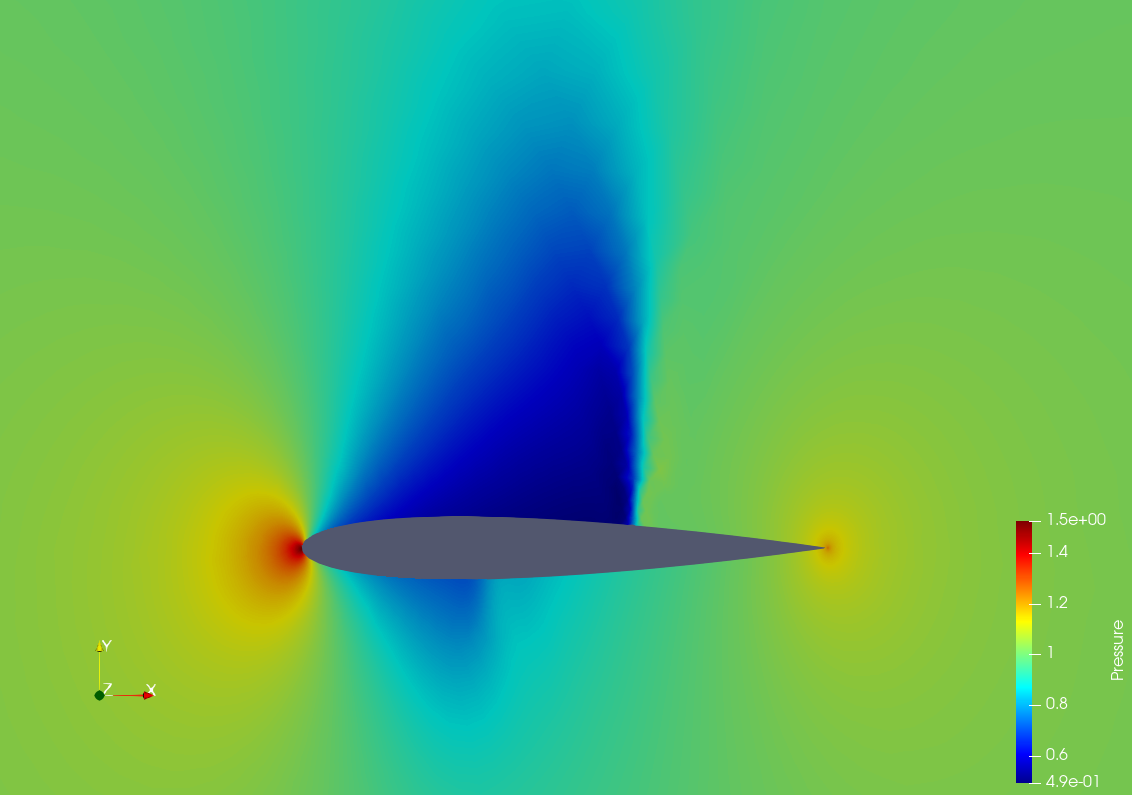}
\caption{ Pressure contours for the baseline NACA 0012 airfoil}
 \label{baseline}
\end{figure}
   
\begin{figure}[htp]
\centering
\includegraphics[width=0.8\textwidth]{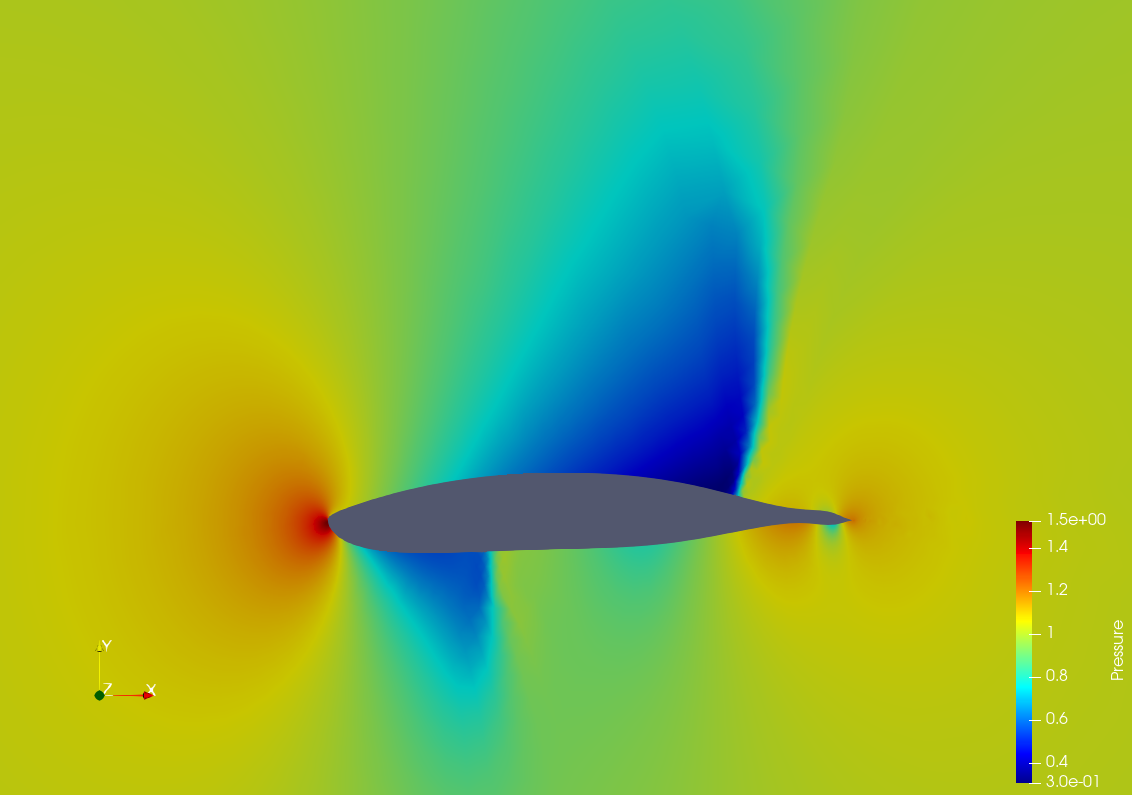}
\caption{Pressure contours for a randomly deformed airfoil} 
\label{random}
\end{figure}
 To demonstrate the performance of SGE-Kriging on a practical enginering application, here we consider an  aerodynamic modeling problem of the NACA 0012 airfoil in the transonic, inviscid flow regime. Our goal is to train different surrogate models to predict the coefficient of drag ($C_d$) for the NACA 0012 airfoil shape. To this end, we modify the airfoil with the Hicks-Henne bump functions \cite{hicks1978wing} by adding perturbations on the upper and lower surfaces of the baseline depicted in Fig. \ref{baseline}, namely
 \begin{equation}
 \begin{cases}
\Lambda_u=\sum_{i=1}^{n}\delta_{ui} f_i(x),\,\ \delta_{ui}\in[-0.01{\rm m},0.01{\rm m}](i=1,\ldots,n) \\
\Lambda_l=\sum_{i=1}^{n}\delta_{li} f_i(x),\,\ \,\ \delta_{li}\in[-0.01{\rm m},0.01{\rm m}],(i=1,\ldots,n)
 \end{cases},
\label{bumpfunction}
\end{equation}
in which $f_i(x)={\rm sin}^3(\pi x^{e(i)})$, and $e(i)={\rm log}(0.5)/{\rm log}(x_i)$. Here $x_i\in[0,1](i=1,\ldots,n)$ indicates the $x$-location of the $i$-th bump, and the corresponding bump amplitudes $\delta_{ui}$ and $\delta_{li}$ are the design parameters.

 In this paper, we use SU2 \cite{economon2016su2}, an open-source multiphysics simulation and design software, to obtain the aerodynamic characteristic of different airfoils, where direct flow analyses and discrete adjoint analyses are implemented by solving Euler and adjoint Euler equations at a freestream Mach number of 0.8 and an angle of attack of 1.25 degrees. The pressure contours of the baseline of NACA 0012 airfoil and a randomly deformed airfoil obtained with SU2 are presented in Fig. \ref{baseline} and Fig. \ref{random}, respectively. As one can see, the airfoil shape has a strong impact on the aerodynamic characteristics.

  \begin{figure}[htp]
  \centering  \includegraphics[width=0.95\textwidth]{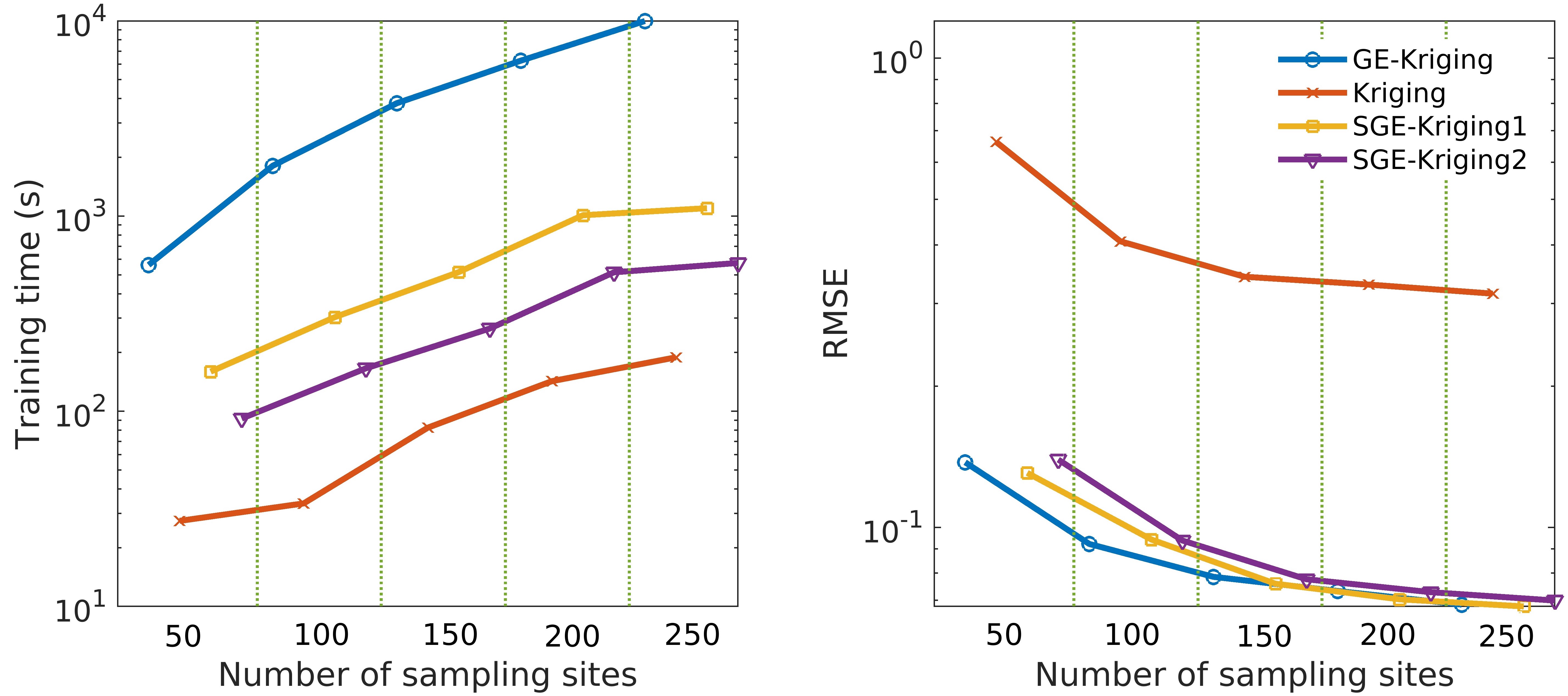}
  \caption{Comparison of the performance of Kriging, GE-Kriging and SGE-Kriging (SGE-Kriging1 = updating the residual terms, SGE-Kriging2 = neglecting the residual terms) with $m=10$ using biquadratic spline correlation function for NACA 0012 airfoil with 38 design parameters}
  \label{NACA_38}
  \end{figure}
  
  In the present work, 19 bumps are uniformly set, namely, $x_1=0.05,x_2=0.1,...,x_{19}=0.95$, on the upper and lower surfaces of the baseline respectively, which results in 38 design parameters. To construct surrogate models, we randomly generate 1000 samples in the parameter space. The first $N(N<1000)$ samples are used to train various surrogate models, and the remaining samples are used to test the model accuracy. Here the training process is performed on the same HPC-cluster534 on the UCloud platform (Inter Xeon Gold 6130 CPU, 32-cores, 2.1GHz). The comparisons of various surrogate models are depicted in Fig \ref{NACA_38}, and the detailed RMSEs and training times are listed in Table \ref{tab:NACA}. It shows that the proposed SGE-Kriging provides nearly consistent prediction accuracy compared to GE-Kriging, but its training cost is reduced sharply, which demonstrates the effectiveness of the proposed method for high-dimensional engineering problems. For the future, it is planned to apply SGE-Kriging to industrial engineering design optimization problems \cite{bouhlel2020scalable}. 
  
   \begin{table}[htbp]
  \centering
  \caption{Comparison of different surrogate models with 250 training samples for the coefficient of drag of NACA 0012 airfoil}
  \begin{tabular}{ccc}
  \toprule
  Model  & RMSE  & Traning time (s)  \\
 \midrule
 Kriging &   0.3148 & 188.87 \\ GE-Kriging &  0.0685 &  9983.20  \\ SGE-Kriging 1 & 0.0678 &  1096.98 \\  SGE-Kriging 2 &   0.0699 &  574.96   \\ 
  \bottomrule
  \end{tabular}
  \label{tab:NACA}
  \end{table}
  
 \section{Conclusions}
 \label{sec:conclusions}
  In this paper, we have proposed a novel variant of GE-Kriging, termed sliced gradient-enhanced Kriging (SGE-Kriging), with the designated goal to increase the efficiency for constructing high-dimensional surrogate models that incorporate gradient information. The contribution of this work is twofold. First, by splitting the sample sites into multiple non-overlapping slices, we propose a family of approximate ``sliced'' likelihood functions based on Bayes' theorem. In this family, we assume the conditional independence  of data samples from slices that are far away from each other. This allows us to describe the correlation of the whole training sample set with multiple small correlation matrices rather than a single big one. Second, we propose to tune the hyper-parameters of SGE-Kriging by learning its relationship with the derivative-based global sensitivity indices.
  
  The validity of the sliced likelihood function is firstly demonstrated with two visualized test functions. The results show that the sliced likelihood function yields a satisfactory approximation of the true one. It is generally observed to feature fewer local minima, but yet it gives optimal or near optimal hyper-parameter values. Then, the performance of SGE-Kriging is also tested 
 with the 8-dimensional Borehole function benchmark problem  and  two high-dimensional analytical functions. The results show that the hyper-parameters of SGE-Kriging can be well determined based on the sensitivity information. The contribution of the residual terms are marginal and can safely be neglected in practical applications. The accuracy of SGE-Kriging is comparable to GE-Kriging, but its training efficiency is improved by one to two orders of magnitude. Finally, SGE-Kriging is applied to an aerodynamic modeling problem of the NACA 0012 airfoil with 38 input parameters. The results confirm again that SGE-Kriging is promising for fitting high-dimensional problems. 
  
  To ensure the robustness of SGE-Kriging, one should split the sample set properly such that the number of sample sites within each slice is not too small, and the slice number can be adjusted dynamically according to the number of sample sites. However, we remark that the optimal number of slices $m$ is a problem-dependent parameter, and our numerical experiment suggests that $m=10$ is a conservative option. For practical application, we suggest setting $m = \lfloor N/5 \rfloor$, so that at least 5 samples are assigned to every slice .  In the future, it is planned to apply the newly developed SGE-Kriging model to a high-dimensional design optimization problem.

\section{Acknowledgement}
  This work is supported by the Austrian Research Promotion Agency FFG under the e!MISSION programme, project ``InduHeat'', project number 881147.
  
\appendix
\section{The biquadratic spline correlation function}
\label{sec:biquadspline}
With $\xi = \xi(|x-\hat{x}|;\theta):=\theta|x-\hat{x}|$, 
the biquadratic spline correlation function is defined as 
 \begin{equation}
     R(|x-\hat{x}|;\theta)=
    \begin{cases}
    1-15\xi^2+35\xi^3-\frac{195}{8}\xi^4,\qquad\qquad 0\leq \xi< 0.4, \\ \frac{5}{3}-\frac{20}{3}\xi+10\xi^2-\frac{20}{3}\xi^3+\frac{5}{3}\xi^4,\qquad 0.4\leq \xi< 1,\\
    0,\qquad\qquad\qquad\qquad\qquad\qquad\qquad\quad 1\leq \xi.
    \end{cases}
    \label{biquadracticfunction}
    \end{equation}
    
   The first-order partial derivative of the biquadratic spline correlation function is
    \begin{eqnarray*}
    \frac {\partial R(\boldsymbol{x},\hat{\boldsymbol{x}};\boldsymbol{\theta})}{\partial {x_k}}=\frac{\partial R(|x_k-\hat{x}_k|;\theta_k)}{\partial {x_k}}\prod_{i=1,i\ne k}^nR(|x_i-\hat{x}_i|;\theta_i),
    \end{eqnarray*}
    where 
   \begin{eqnarray*}
    \frac{\partial{R(|x-\hat{x}|;\theta)}}{\partial x} =
    \begin{cases}
    \left(-30\xi+105\xi^2-\frac{195}{2}\xi^3\right)\theta \rm{sign}|\emph x-\hat{\emph x}|,\qquad 0\leq \xi< 0.4, \\ \left(-\frac{20}{3}+20\xi-20\xi^2+\frac{20}{3}\xi^3\right)\theta \rm{sign}|\emph x-\hat{\emph x}|,\;\ 0.4\leq \xi< 1,\\
    0,\qquad\qquad\qquad\qquad\qquad\qquad\qquad\qquad\qquad 1 \leq \xi.
    \end{cases}
    \end{eqnarray*}
    
    Similarly, the second-order partial derivative of the biquadratic spline correlation function are
    \begin{eqnarray*}
    \frac {\partial^2 R(\boldsymbol{x},\hat{\boldsymbol{x}};\boldsymbol{\theta})}{\partial {x_{k}}\partial {\hat x_{l}}}
    =
    \begin{cases}
    \frac{\partial^2 R(|x_{k}-\hat{x}_{k}|;\theta_{k})}{\partial {x_{k}} \partial \hat{x}_{k}}\prod_{i=1,i\ne k}^nR(|x_i-\hat{x}_i|;\theta_i), \qquad\qquad\quad \,\ k= l, \\ \frac{\partial R(|x_{k}-\hat{x}_{k}|;\theta_{k})}{\partial {x_{k}}}\frac{\partial R(|x_{l}-\hat{x}_{l}|;\theta_{l})}{\partial\hat {x}_{l}}\prod_{i=1,i\ne k,l}^nR(|x_i-\hat{x_i}|;\theta_i),\;\ k\neq l.
    \end{cases}
    \end{eqnarray*}
    where
    \begin{eqnarray*}
    \frac{\partial^2{R(|x-\hat{x}|;\theta)}}{\partial x \partial \hat x} =
    \begin{cases}
    -\left(-30+210\xi-\frac{585}{2}\xi^2\right)\theta^2,\qquad 0\leq \xi< 0.4, \\ -\left(20-40\xi+20\xi^2\right)\theta^2,\;\ \qquad\quad 0.4\leq \xi< 1,\\
    0,\qquad\qquad\qquad\qquad\qquad\qquad\qquad 1\leq \xi.
    \end{cases}
    \end{eqnarray*}

%% file: SISC_main.bbl
\begin{thebibliography}{10}

\bibitem{abiodun2018state}
{\sc O.~I. Abiodun, A.~Jantan, A.~E. Omolara, K.~V. Dada, N.~A. Mohamed, and
  H.~Arshad}, {\em State-of-the-art in artificial neural network applications:
  A survey}, Heliyon, 4 (2018), p.~e00938.

\bibitem{anderson1999aerodynamic}
{\sc W.~K. Anderson and V.~Venkatakrishnan}, {\em Aerodynamic design
  optimization on unstructured grids with a continuous adjoint formulation},
  Computers \& Fluids, 28 (1999), pp.~443--480.

\bibitem{blatman2011adaptive}
{\sc G.~Blatman and B.~Sudret}, {\em Adaptive sparse polynomial chaos expansion
  based on least angle regression}, Journal of computational Physics, 230
  (2011), pp.~2345--2367.

\bibitem{bouhlel2020scalable}
{\sc M.~A. Bouhlel, S.~He, and J.~R. Martins}, {\em Scalable gradient--enhanced
  artificial neural networks for airfoil shape design in the subsonic and
  transonic regimes}, Structural and Multidisciplinary Optimization, 61 (2020),
  pp.~1363--1376.

\bibitem{bouhlel2019gradient}
{\sc M.~A. Bouhlel and J.~R. Martins}, {\em Gradient-enhanced kriging for
  high-dimensional problems}, Engineering with Computers, 35 (2019),
  pp.~157--173.

\bibitem{chen2019screening}
{\sc L.~Chen, H.~Qiu, L.~Gao, C.~Jiang, and Z.~Yang}, {\em A screening-based
  gradient-enhanced kriging modeling method for high-dimensional problems},
  Applied Mathematical Modelling, 69 (2019), pp.~15--31.

\bibitem{chen2020optimization}
{\sc L.~Chen, H.~Qiu, L.~Gao, C.~Jiang, and Z.~Yang}, {\em Optimization of
  expensive black-box problems via gradient-enhanced kriging}, Computer Methods
  in Applied Mechanics and Engineering, 362 (2020), p.~112861.

\bibitem{cheng2018adaptive}
{\sc K.~Cheng and Z.~Lu}, {\em Adaptive sparse polynomial chaos expansions for
  global sensitivity analysis based on support vector regression}, Computers \&
  Structures, 194 (2018), pp.~86--96.

\bibitem{cheng2021adaptive}
{\sc K.~Cheng and Z.~Lu}, {\em Adaptive bayesian support vector regression
  model for structural reliability analysis}, Reliability Engineering \& System
  Safety, 206 (2021), p.~107286.

\bibitem{cheng2020surrogate}
{\sc K.~Cheng, Z.~Lu, C.~Ling, and S.~Zhou}, {\em Surrogate-assisted global
  sensitivity analysis: an overview}, Structural and Multidisciplinary
  Optimization, 61 (2020), pp.~1187--1213.

\bibitem{chung2002using}
{\sc H.-S. Chung and J.~J. Alonso}, {\em Using gradients to construct cokriging
  approximation models for high-dimensional design optimization problems}, in
  40th AIAA Aerospace Sciences Meeting \& Exhibit, 2002, p.~317.

\bibitem{da2015global}
{\sc S.~Da~Veiga}, {\em Global sensitivity analysis with dependence measures},
  Journal of Statistical Computation and Simulation, 85 (2015), pp.~1283--1305.

\bibitem{economon2016su2}
{\sc T.~D. Economon, F.~Palacios, S.~R. Copeland, T.~W. Lukaczyk, and J.~J.
  Alonso}, {\em Su2: An open-source suite for multiphysics simulation and
  design}, Aiaa Journal, 54 (2016), pp.~828--846.

\bibitem{fu2020distance}
{\sc C.~Fu, P.~Wang, L.~Zhao, and X.~Wang}, {\em A distance correlation-based
  kriging modeling method for high-dimensional problems}, Knowledge-Based
  Systems, 206 (2020), p.~106356.

\bibitem{han2013improving}
{\sc Z.-H. Han, S.~G{\"o}rtz, and R.~Zimmermann}, {\em Improving
  variable-fidelity surrogate modeling via gradient-enhanced kriging and a
  generalized hybrid bridge function}, Aerospace Science and technology, 25
  (2013), pp.~177--189.

\bibitem{han2017weighted}
{\sc Z.-H. Han, Y.~Zhang, C.-X. Song, and K.-S. Zhang}, {\em Weighted
  gradient-enhanced kriging for high-dimensional surrogate modeling and design
  optimization}, Aiaa Journal, 55 (2017), pp.~4330--4346.

\bibitem{hicks1978wing}
{\sc R.~M. Hicks and P.~A. Henne}, {\em Wing design by numerical optimization},
  Journal of Aircraft, 15 (1978), pp.~407--412.

\bibitem{sobol2010derivative}
{\sc S.~K. Ilya M.~Sobol'}, {\em Derivative based global sensitivity measures},
  Procedia-Social and Behavioral Sciences, 2 (2010), pp.~7745--7746.

\bibitem{jones1998efficient}
{\sc D.~R. Jones, M.~Schonlau, and W.~J. Welch}, {\em Efficient global
  optimization of expensive black-box functions}, Journal of Global
  optimization, 13 (1998), pp.~455--492.

\bibitem{Koehler96computerexperiments}
{\sc J.~R. Koehler and A.~B. Owen}, {\em 9 computer experiments}, in Design and
  analysis of experiments, S.~Ghosh and C.~R. Rao, eds., vol.~13 of Handbook of
  statistics, North-Holland, 1996, pp.~261--308,
  \href{http://dx.doi.org/10.1016/s0169-7161(96)13011-x}{doi:\nolinkurl{10.1016/s0169-7161(96)13011-x}},
  \url{https://doi.org/10.1016/s0169-7161(96)13011-x}.

\bibitem{kowalik1968methods}
{\sc J.~S. Kowalik and M.~R. Osborne}, {\em Methods for unconstrained
  optimization problems}, vol.~13, Elsevier Publishing Company, 1968.

\bibitem{krige1951statistical}
{\sc D.~G. Krige}, {\em A statistical approach to some basic mine valuation
  problems on the witwatersrand}, Journal of the Southern African Institute of
  Mining and Metallurgy, 52 (1951), pp.~119--139.

\bibitem{kucherenko2016derivative}
{\sc S.~Kucherenko and S.~Song}, {\em Derivative-based global sensitivity
  measures and their link with sobol’sensitivity indices}, in Monte Carlo and
  Quasi-Monte Carlo Methods, Springer, 2016, pp.~455--469.

\bibitem{le2013multi}
{\sc L.~Le~Gratiet}, {\em Multi-fidelity Gaussian process regression for
  computer experiments}, PhD thesis, Universit{\'e} Paris-Diderot-Paris VII,
  2013.

\bibitem{liu2003development}
{\sc W.~Liu}, {\em Development of gradient-enhanced kriging approximations for
  multidisciplinary design optimization}, University of Notre Dame, 2003.

\bibitem{liu2002gradient}
{\sc W.~Liu and S.~Batill}, {\em Gradient-enhanced response surface
  approximations using kriging models}, in 9th AIAA/ISSMO symposium on
  multidisciplinary analysis and optimization, 2002, p.~5456.

\bibitem{lophaven2002dace}
{\sc S.~N. Lophaven, H.~B. Nielsen, J.~S{\o}ndergaard, et~al.}, {\em DACE: a
  Matlab kriging toolbox}, vol.~2, Citeseer, 2002.

\bibitem{doi:10.1137/21M1397908}
{\sc L.~Lu, R.~Pestourie, W.~Yao, Z.~Wang, F.~Verdugo, and S.~G. Johnson}, {\em
  Physics-informed neural networks with hard constraints for inverse design},
  SIAM Journal on Scientific Computing, 43 (2021), pp.~B1105--B1132,
  \href{http://dx.doi.org/10.1137/21M1397908}{doi:\nolinkurl{10.1137/21M1397908}}.

\bibitem{moon2012two}
{\sc H.~Moon, A.~M. Dean, and T.~J. Santner}, {\em Two-stage sensitivity-based
  group screening in computer experiments}, Technometrics, 54 (2012),
  pp.~376--387.

\bibitem{morris1993bayesian}
{\sc M.~D. Morris, T.~J. Mitchell, and D.~Ylvisaker}, {\em Bayesian design and
  analysis of computer experiments: use of derivatives in surface prediction},
  Technometrics, 35 (1993), pp.~243--255.

\bibitem{neidinger2010introduction}
{\sc R.~D. Neidinger}, {\em Introduction to automatic differentiation and
  matlab object-oriented programming}, SIAM review, 52 (2010), pp.~545--563.

\bibitem{queipo2005surrogate}
{\sc N.~V. Queipo, R.~T. Haftka, W.~Shyy, T.~Goel, R.~Vaidyanathan, and P.~K.
  Tucker}, {\em Surrogate-based analysis and optimization}, Progress in
  aerospace sciences, 41 (2005), pp.~1--28.

\bibitem{rasmussen2003gaussian}
{\sc C.~E. Rasmussen}, {\em Gaussian processes in machine learning}, in Summer
  school on machine learning, Springer, 2003, pp.~63--71.

\bibitem{rosenbaum2013efficient}
{\sc B.~Rosenbaum}, {\em Efficient global surrogate models for responses of
  expensive simulations},  (2013).

\bibitem{sacks1989design}
{\sc J.~Sacks, W.~J. Welch, T.~J. Mitchell, and H.~P. Wynn}, {\em Design and
  analysis of computer experiments}, Statistical science, 4 (1989),
  pp.~409--423.

\bibitem{saltelli2008global}
{\sc A.~Saltelli, M.~Ratto, T.~Andres, F.~Campolongo, J.~Cariboni, D.~Gatelli,
  M.~Saisana, and S.~Tarantola}, {\em Global sensitivity analysis: the primer},
  John Wiley \& Sons, 2008.

\bibitem{smola2004tutorial}
{\sc A.~J. Smola and B.~Sch{\"o}lkopf}, {\em A tutorial on support vector
  regression}, Statistics and computing, 14 (2004), pp.~199--222.

\bibitem{vapnik1999nature}
{\sc V.~Vapnik}, {\em The nature of statistical learning theory}, Springer
  science \& business media, 1999.

\bibitem{wei2015variable}
{\sc P.~Wei, Z.~Lu, and J.~Song}, {\em Variable importance analysis: a
  comprehensive review}, Reliability Engineering \& System Safety, 142 (2015),
  pp.~399--432.

\bibitem{williams2006gaussian}
{\sc C.~K. Williams and C.~E. Rasmussen}, {\em Gaussian processes for machine
  learning}, vol.~2, MIT press Cambridge, MA, 2006.

\bibitem{xiu2002wiener}
{\sc D.~Xiu and G.~E. Karniadakis}, {\em The wiener--askey polynomial chaos for
  stochastic differential equations}, SIAM journal on scientific computing, 24
  (2002), pp.~619--644.

\bibitem{doi:10.1137/21M1432739}
{\sc R.~Zhang, S.~Mak, and D.~Dunson}, {\em Gaussian process subspace
  prediction for model reduction}, SIAM Journal on Scientific Computing, 44
  (2022), pp.~A1428--A1449,
  \href{http://dx.doi.org/10.1137/21M1432739}{doi:\nolinkurl{10.1137/21M1432739}},
  \url{https://doi.org/10.1137/21M1432739}.

\bibitem{zhao2020efficient}
{\sc L.~Zhao, P.~Wang, B.~Song, X.~Wang, and H.~Dong}, {\em An efficient
  kriging modeling method for high-dimensional design problems based on maximal
  information coefficient}, Structural and Multidisciplinary Optimization, 61
  (2020), pp.~39--57.

\bibitem{zimmermann2013maximum}
{\sc R.~Zimmermann}, {\em On the maximum likelihood training of
  gradient-enhanced spatial gaussian processes}, SIAM Journal on Scientific
  Computing, 35 (2013), pp.~A2554--A2574.

\bibitem{zimmermann2015}
{\sc R.~Zimmermann}, {\em On the condition number anomaly of {G}aussian
  correlation matrices}, Linear Algebra and its Applications, 466 (2015),
  pp.~512--526,
  \href{http://dx.doi.org/10.1016/j.laa.2014.10.038}{doi:\nolinkurl{10.1016/j.laa.2014.10.038}},
  \url{http://dx.doi.org/10.1016/j.laa.2014.10.038}.

\end{thebibliography}
